\documentclass[pmlr,twocolumn,10pt]{jmlr} % W&CP article

\mlhtrack{proceedings}

\newif\iffinal
\finaltrue  % toggle this flag for camera-ready version after acceptance

\iffinal
    \ifmlhneedspmlr
      \jmlrvolume{}
      \jmlryear{}
    \fi
    \ifmlhfindings \jmlrproceedings{}{}\fi
    \ifmlhdemo     \jmlrproceedings{}{}\fi
    \jmlrworkshop{}
\else
    \jmlrproceedings{}{}
    \jmlrworkshop{}
\fi

\usepackage{booktabs}
\usepackage{siunitx}
\usepackage{graphicx}
\usepackage{subcaption}

\usepackage[switch]{lineno}

\theorembodyfont{\upshape}
\theoremheaderfont{\scshape}
\theorempostheader{:}
\theoremsep{\newline}

\title[What to Preserve, Where to Adapt?]
{What to Preserve, Where to Adapt:\\
Understanding Forgetting in Continual Gynecological Image Segmentation}

\author{%
\Name{Amal Saqib} \Email{amal.saqib@mbzuai.ac.ae}\\
\Name{Tausifa Jan Saleem} \Email{tausifa.saleem@mbzuai.ac.ae}\\
\Name{Numan Saeed} \Email{numan.saeed@mbzuai.ac.ae}\\
\Name{Mohammad Yaqub} \Email{mohammad.yaqub@mbzuai.ac.ae}\\
\addr Mohamed bin Zayed University of Artificial Intelligence (MBZUAI), Abu Dhabi, UAE
}

\begin{document}

\maketitle

\ifmlhdemo\else

%===========================================================================================================================================================================================%
\begin{abstract}
The clinical management of gynecological diseases often relies on medical
imaging for diagnosis, treatment planning, and follow-up. Segmentation in this
setting is challenging because successive tasks may differ in imaging
modality, target anatomy, pathology, and annotation structure. Continual learning allows models to adapt to new tasks without simultaneous
access to previous datasets. However, when successive tasks differ
substantially, learning a new task can degrade performance on earlier ones,
a problem known as catastrophic forgetting. Understanding where adaptation disrupts previous
knowledge can help guide the design of more targeted continual-learning
strategies. We investigate how forgetting changes as different parts of an
encoder--decoder network are allowed to adapt. We progressively expand the
trainable region of a 3D nnU-Net backbone from the bottleneck toward input- and output-proximal blocks. Under a shared learning rate, adaptation near the bottleneck largely
preserves previous-task performance but provides limited current-task
learning, whereas broader adaptation improves current-task performance but
sharply increases forgetting. This trade-off persists even when the average
change in trainable backbone parameters is approximately comparable. Assigning
different learning rates to different blocks substantially reduces forgetting
when part of the backbone is trainable, although this changes both the size
and location of the updates. Forgetting still increases as more blocks are
trained and remains severe when the full backbone is updated. These results
show that forgetting depends not only on how much the model changes, but also
on which parts of the model are allowed to change.
\end{abstract}


%===========================================================================================================================================================================================%

\fi

\ifmlhneedsstatements

%===========================================================================================================================================================================================%
\paragraph*{Data and Code Availability}
This study uses three publicly available datasets: the Uterine Myoma MRI
Dataset (UMD)~\cite{umd}, the Endometrial Cancer PET/CT Image Dataset for
Semantic Segmentation and Detection of Hypermetabolic Regions
(ECPC-IDS)~\cite{ecpc}, and the UTHealth Endometriosis MRI Dataset
(UT-EndoMRI)~\cite{endomri}. The datasets are available from their original
sources and remain subject to their respective access and usage conditions.
Code for dataset preparation, model training, continual-learning experiments,
and analysis are available at
\url{https://github.com/AmalSaqib/What-to-Preserve-Where-to-Adapt}.

\paragraph*{Institutional Review Board (IRB)}
This research does not require IRB approval.

\fi

%===========================================================================================================================================================================================%

\section{Introduction}
\label{sec:introduction}

Medical image segmentation models are commonly developed under the assumption
that all training data are available simultaneously. In clinical practice,
however, imaging datasets and segmentation objectives may become available
sequentially across institutions or studies. Retaining and combining earlier
datasets may also be constrained by privacy, security, storage, and
institutional data-governance requirements. Continual learning therefore aims
to extend an existing model using only current-task data while preserving its
previously learned capabilities. This setting is particularly relevant to gynecological imaging, where
segmentation supports diagnosis, treatment planning, and longitudinal
assessment, but tasks can differ substantially in imaging modality, target
anatomy, pathology, and annotation structure
~\cite{kinkel2006diagnosis,sala2013added,umd,endomri,ecpc}. For example, a
model may need to adapt from segmenting uterine anatomy in MRI to segmenting
tumors in PET/CT or endometriosis-related structures in another MRI dataset.
These differences provide a clinically relevant setting for studying
adaptation under heterogeneous and sequentially available data. Sequential adaptation can cause catastrophic forgetting, in which learning a
new task degrades performance on earlier tasks
~\cite{catastrophic,cl_medical_survey,css_survey}. Existing approaches
mitigate forgetting through replay, regularization, knowledge distillation,
or parameter isolation~\cite{ewc,lwf,gem,agem}. However, they do not fully
explain how the location and extent of adaptation within the network affect
the trade-off between retaining previous tasks and learning the current task.

U-Net and its extensions, including nnU-Net, are widely used for medical image
segmentation~\cite{unet,nnunet}. Their encoder--decoder structure is not
uniform across depth: the encoder, bottleneck, decoder, and skip connections
perform different roles in producing a segmentation
~\cite{lee_unet_information,yang2025pruning,hassler2025lean}. Allowing
different regions to adapt may therefore affect previous and current tasks
differently. A further complication is that a shared learning rate need not
update all layers equally because their gradient scales can differ
~\cite{tang_layerwise_cl,kumar2024fine}. Consequently, differences attributed
to adaptation depth may reflect both which regions are trainable and how
strongly they are updated. Restricting adaptation too strongly may prevent
learning a new clinical task, whereas allowing excessive change may erase
previous capabilities. Understanding this behavior can guide methods toward a
more effective balance between retention and adaptation.

We investigate these factors in continual gynecological image segmentation.
Starting from the same first-task checkpoint, we progressively expand the
trainable region of a 3D nnU-Net backbone from the bottleneck toward
input- and output-proximal blocks. We measure previous-task forgetting,
current-task performance, and aggregate and block-level parameter
displacement. We first compare three adaptation extents with approximately
comparable average changes in their trainable backbone parameters. This tests
whether expanding the trainable region produces more forgetting simply
because more parameter movement occurs. We then compare a shared learning
rate with static block-specific rates calibrated from initial relative
gradient norms over identical trainable regions. Because calibration changes
both overall update magnitude and its distribution, we interpret it as an
analysis of update scaling rather than an isolation of either factor. We also
examine an MRI-to-MRI transition to determine whether the pattern is limited
to the larger MRI-to-PET/CT shift.

Our main contributions are:
\begin{itemize}
\item We present a controlled evaluation framework that combines nested
trainable regions with aggregate and block-level parameter-displacement
analysis to study continual adaptation across encoder--decoder depth.

\item By approximately matching average per-parameter movement across selected
adaptation extents, we show that similar parameter displacement can produce
substantially different retention and current-task performance.

\item Using gradient-calibrated block-specific learning rates, we show that
update scaling can substantially improve retention at intermediate adaptation
extents. Because calibration changes both update magnitude and its
distribution, the results indicate that both where and how strongly a network
adapts must be considered.
\end{itemize}
%===========================================================================================================================================================================================%

\section{Related Work}
\label{sec:related_work}

\paragraph{Continual Learning for Medical Image Segmentation.}
Methods for mitigating catastrophic forgetting include regularization,
knowledge distillation, replay, and parameter isolation
~\cite{ewc,zenke2017si,lwf,gem,agem,cl_medical_survey}. In medical image
segmentation, these approaches have been studied across changes in imaging
domain, anatomy, and segmentation objective
~\cite{gonzalez2020continual,ranem2022hippocampus}. This literature primarily
compares strategies for balancing retention and current-task learning. Our
work instead examines how that balance changes as different regions of an
encoder--decoder backbone are adapted.

\paragraph{Adaptation Across Network Depth.}
Continual-learning studies show that representational change and parameter
sensitivity vary across network depth
~\cite{ramasesh2020anatomy,zhao2023parameters}. Selective freezing further
shows that the choice of trainable layers affects retention
~\cite{sorrenti2023selective}, while fine-tuning studies find that useful
adaptation regions depend on the distribution shift
~\cite{lee2022surgical}. Information-flow and pruning analyses of
encoder--decoder networks also report different contributions across the
encoder, bottleneck, decoder, and skip connections
~\cite{lee_unet_information,yang2025pruning,hassler2025lean}. However, these
studies do not separate the effect of adaptation extent from differences in
update magnitude across the selected regions.

\paragraph{Layer-Wise Update Scale.}
A shared learning rate can produce unequal updates because gradient magnitudes
and parameter scales vary between layers
~\cite{tang_layerwise_cl,kumar2024fine}. Layer-wise optimization addresses
such differences, but update scale and adaptation extent are typically
examined separately. We connect them by comparing adaptation extents at
approximately matched average parameter displacement and by evaluating shared
and block-specific learning rates over identical trainable regions.
%===========================================================================================================================================================================================%
\begin{figure*}[t]
\centering
\includegraphics[width=0.84\textwidth]{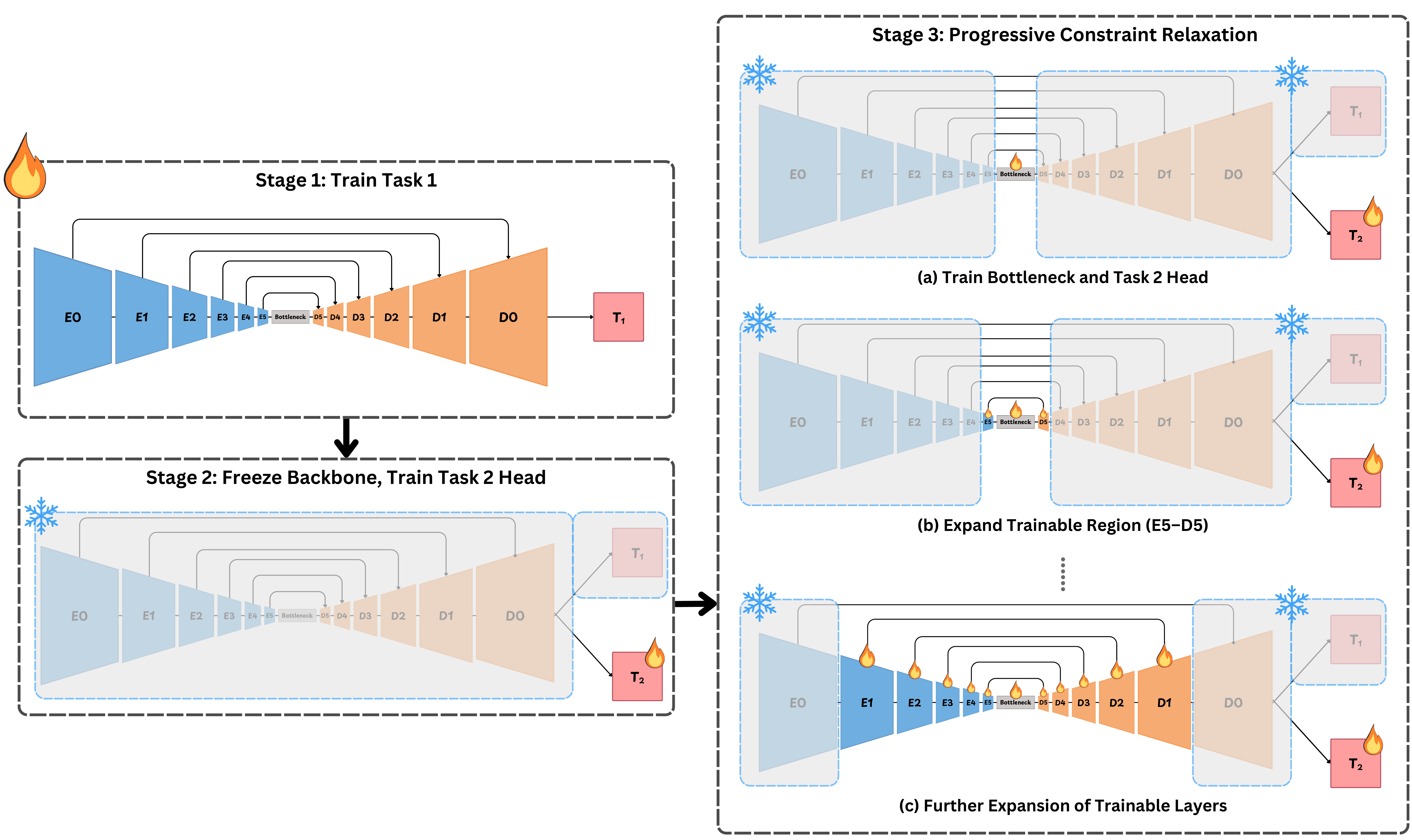}
\caption{Overview of the depth-constrained adaptation analysis. Each
configuration is initialized from the same Task-1 checkpoint. During Task-2
adaptation, the trainable region is progressively expanded from the
bottleneck toward input- and output-proximal blocks. Pre- and post-adaptation
parameters are compared to measure movement across the encoder--decoder
backbone.}
\label{fig:method}
\end{figure*}

\section{Method}
\label{sec:method}

We examine how continual forgetting changes with the extent of backbone
adaptation and the distribution of parameter updates across network depth.
We progressively expand the trainable region, measure aggregate and
block-level parameter displacement, and compare shared and controlled update
scales.

\subsection{Encoder--Decoder Decomposition}
\label{sec:encoder_decoder_decomposition}

We use a shared 3D nnU-Net backbone with task-specific segmentation heads. The
backbone contains encoder blocks \(E_0,\ldots,E_5\), a bottleneck \(B\), and
decoder blocks \(D_5,\ldots,D_0\). Blocks \(E_0\) and \(D_0\) are closest to
the input and output, respectively, whereas \(E_5\), \(B\), and \(D_5\) form
the deepest region. Each block contains its convolutional layers and
corresponding downsampling or upsampling operation. Encoder and decoder blocks
at matching resolutions are treated as pairs, as shown in
Figure~\ref{fig:method}.

For concise notation, we define
\[
\mathcal{R}_k
=
\{E_k,\ldots,E_5,B,D_5,\ldots,D_k\},
\qquad k\in\{0,\ldots,5\}.
\]
Here, \(\mathcal{R}_0\) contains the full backbone, and increasing \(k\)
progressively restricts adaptation toward the bottleneck. We additionally
evaluate \(B\) alone as bottleneck-only adaptation.

\subsection{Depth-Constrained Adaptation}
\label{sec:depth_constrained_adaptation}

Each configuration is initialized independently from the same Task-1
checkpoint. During Task-2 training, the current-task head and selected
backbone blocks are trainable; the remaining backbone and previous-task head
are frozen. Starting with \(B\), the trainable region is expanded by adding
matched encoder--decoder pairs:
\[
B,\ \mathcal{R}_5,\ \mathcal{R}_4,\ldots,\mathcal{R}_0.
\]
These nested configurations allow retention and current-task performance to
be compared as adaptation extent increases.

\subsection{Parameter Displacement and Update Allocation}
\label{sec:parameter_displacement}

We measure how much the trainable shared-backbone parameters change between
the start and end of Task-2 adaptation. Let \(\mathcal{S}\) be the set of
trainable scalar backbone parameters and \(N=|\mathcal{S}|\). For parameter
\(i\), \(\theta_i^{\mathrm{pre}}\) and \(\theta_i^{\mathrm{post}}\) denote its
values before and after adaptation, respectively. The root-mean-square (RMS)
parameter displacement is
\[
D_{\mathrm{RMS}}
=
\sqrt{
\frac{1}{N}
\sum_{i\in\mathcal{S}}
\left(\theta_i^{\mathrm{post}}-\theta_i^{\mathrm{pre}}\right)^2
}.
\]
This measures the average magnitude of change per trainable parameter.
Dividing by \(N\) allows configurations with different numbers of trainable
parameters to be compared.

Let \(\boldsymbol{\theta}_{\mathcal{S}}^{\mathrm{pre}}\) and
\(\boldsymbol{\theta}_{\mathcal{S}}^{\mathrm{post}}\) be vectors containing
all parameters in \(\mathcal{S}\). Relative displacement is
\[
D_{\mathrm{rel}}
=
\frac{
\left\|
\boldsymbol{\theta}_{\mathcal{S}}^{\mathrm{post}}
-
\boldsymbol{\theta}_{\mathcal{S}}^{\mathrm{pre}}
\right\|_2
}{
\left\|\boldsymbol{\theta}_{\mathcal{S}}^{\mathrm{pre}}\right\|_2+\epsilon
}
\times100,
\]
where \(\epsilon>0\) prevents division by zero. This expresses the total
parameter change as a percentage of the initial parameter norm. Task-specific
heads are excluded from both measures.

To determine where movement occurs, let
\[
\Delta\boldsymbol{\theta}_b
=
\boldsymbol{\theta}_b^{\mathrm{post}}
-
\boldsymbol{\theta}_b^{\mathrm{pre}}
\]
be the parameter change in trainable block \(b\). Its share of total squared
backbone displacement is
\[
q_b
=
\frac{
\left\|\Delta\boldsymbol{\theta}_b\right\|_2^2
}{
\sum_{c\in\mathcal{B}_{\mathcal{S}}}
\left\|\Delta\boldsymbol{\theta}_c\right\|_2^2
}
\times100,
\]
where \(\mathcal{B}_{\mathcal{S}}\) is the set of trainable backbone blocks.
Thus, \(q_b\) is the percentage of squared parameter movement occurring in
block \(b\), and the shares sum to \(100\%\). Block-level RMS and relative
displacement use the same definitions restricted to one block.

\subsection{Approximate Final-RMS Control}
\label{sec:rms_control}
To test whether average parameter movement alone explains forgetting, we
compare selected adaptation extents with approximately comparable final
\(D_{\mathrm{RMS}}\). Bottleneck-only adaptation (\(B\)) defines the target displacement.
The initial learning rate for \(\mathcal{R}_1\) is then reduced to bring its
final \(D_{\mathrm{RMS}}\) into the same range; \(\mathcal{R}_4\) already
falls within this range at the original learning rate. Similarity is assessed at the final checkpoint rather
than enforced during training. This control does not equalize total
\(\ell_2\) displacement or movement within individual blocks. The learning-rate
calculation is provided in Appendix~\ref{sec:app_rms_targeting}.

\subsection{Scale-Aware Block-Specific Learning Rates}
\label{sec:scale_aware_updates}
Under a shared learning rate, blocks with larger gradients relative to their
parameter scale may receive stronger effective updates. To reduce this
imbalance, we assign each trainable block a fixed learning-rate multiplier
using gradients measured before adaptation.

For block \(b\), we compute its median relative gradient norm over \(K\)
current-task batches:
\[
\bar r_b
=
\operatorname*{median}_{k=1,\ldots,K}
\frac{
\|\nabla_{\boldsymbol{\theta}_b}\mathcal{L}_k\|_2
}{
\|\boldsymbol{\theta}_b\|_2+\epsilon
}.
\]
Here, \(\mathcal{L}_k\) is the current-task loss on batch \(k\), and
\(\boldsymbol{\theta}_b\) contains the parameters of block \(b\). Using the
bottleneck \(B\) as the reference, the multiplier for block \(b\) is
\[
\alpha_b
=
\operatorname{clip}
\left(
\frac{\bar r_B}{\bar r_b},
0.1,10
\right).
\]
Thus, blocks with larger relative gradients receive smaller learning rates,
while blocks with smaller relative gradients receive larger learning rates.
The learning rate for block \(b\) at epoch \(e\) is
\[
\eta_b(e)=\alpha_b\eta(e),
\]
where \(\eta(e)\) is the standard nnU-Net polynomial schedule. Calibration uses \(K=20\) batches without optimizer updates. The multipliers
remain fixed during adaptation; the bottleneck has \(\alpha_B=1\), and the
current-task head uses \(\alpha=1\). This procedure changes both the overall
magnitude and the depth-wise distribution of parameter updates; it does not
match final parameter displacement. The resulting multipliers are reported in
Appendix~\ref{sec:app_lr_multipliers}.

%================================================================================%

\section{Experimental Setup}
\label{sec:experiments}

\subsection{Datasets}
\label{sec:datasets}

We use three public gynecological imaging datasets. UMD~\cite{umd} contains
T2-weighted MRI with labels for uterine wall (UW), uterine cavity (UC), myoma
(MY), and nabothian cyst (NC). ECPC-IDS~\cite{ecpc} contains paired PET/CT
volumes with binary tumor labels. UT-EndoMRI~\cite{endomri} contains
multi-sequence pelvic MRI with labels for uterus (UT), ovaries (OV),
endometriomas (EM), and cysts (CY). For UT-EndoMRI, we use the 103 available T1-weighted fat-suppressed (T1FS)
volumes. T1FS was chosen through an exploratory sequence comparison that
included the held-out evaluation results. We therefore treat the
UMD \(\rightarrow\) UT-EndoMRI experiment as a secondary analysis rather
than an independent confirmatory test. The sequence comparison is provided in
Appendix~\ref{sec:app_endomri_selection}.

\begin{figure}[t]
\floatconts
  {fig:dataset_examples}
  {\caption{Representative cases. (a) UMD: uterine wall (green), uterine
  cavity (cyan), myoma (yellow), and nabothian cyst (magenta).
  (b) ECPC-IDS: tumor annotation (green) over fused PET/CT.
  (c) UT-EndoMRI: uterus (green) and endometrioma (yellow).}}
  {%
    \subfigure[]{%
      \includegraphics[width=0.31\linewidth]{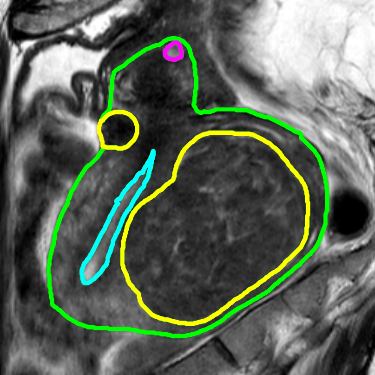}}%
    \hfill
    \subfigure[]{%
      \includegraphics[width=0.31\linewidth]{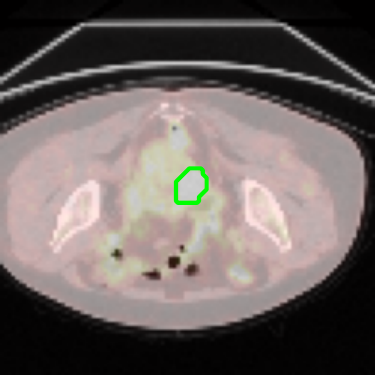}}%
    \hfill
    \subfigure[]{%
      \includegraphics[width=0.31\linewidth]{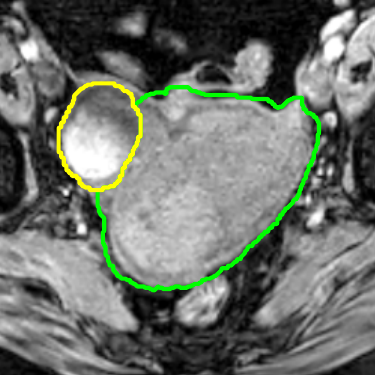}}%
  }
\end{figure}

\begin{table}[t]
\floatconts
  {tab:dataset_summary}
  {\caption{Datasets and case-level splits.}}
  {%
    \small
    \setlength{\tabcolsep}{3pt}
    \renewcommand{\arraystretch}{1.1}
    \begin{tabular*}{\columnwidth}
      {@{\extracolsep{\fill}}lcccc}
    \toprule
    \textbf{Dataset} & \textbf{Modality} &
    \textbf{Train} & \textbf{Eval.} & \textbf{Task} \\
    \midrule
    UMD        & T2-MRI    & 214 & 52 & Multi-class \\
    ECPC-IDS   & PET+CT    & 108 & 26 & Binary \\
    UT-EndoMRI & T1FS-MRI  & 83  & 20 & Multi-class \\
    \bottomrule
    \end{tabular*}
  }
\end{table}

The approximately 80/20 training and evaluation splits contain disjoint case
identifiers and are fixed across methods. Each dataset is processed separately
using the standard nnU-Net pipeline~\cite{nnunet}. UMD and UT-EndoMRI use MRI
and a zero-filled auxiliary channel, whereas ECPC-IDS uses CT and PET.
Additional statistics are reported in
Appendix~\ref{sec:app_dataset_stats}. Each dataset defines a separate task. Baselines use the sequence
UMD \(\rightarrow\) ECPC-IDS \(\rightarrow\) UT-EndoMRI
(Appendix~\ref{sec:app_cl_baselines}). The primary analysis uses
UMD \(\rightarrow\) ECPC-IDS, while UMD \(\rightarrow\) UT-EndoMRI is the
secondary MRI-to-MRI analysis.

\subsection{Implementation Details}
\label{sec:implementation}

Experiments use Lifelong nnU-Net~\cite{lifelong_nnunet} with its automatically
configured 3D full-resolution architecture. Models are trained for 100 epochs
using SGD with momentum \(0.99\), Nesterov acceleration, weight decay
\(3\times10^{-5}\), combined Dice and cross-entropy loss, and the standard
nnU-Net polynomial learning-rate schedule. Each controlled run loads the same UMD weights and creates a new optimizer
after applying its freezing policy; UMD optimizer and momentum states are not
transferred. During periodic validation, the controlled implementation
reconstructs the optimizer and resets momentum identically across controlled
configurations.

For UMD \(\rightarrow\) ECPC-IDS, the shared-LR analysis progressively
expands the trainable region from \(B\) to the full backbone using initial LR
\(0.01\). The approximate-RMS control compares \(B\), \(\mathcal R_4\), and
\(\mathcal R_1\); \(B\) and \(\mathcal R_4\) use LR \(0.01\), while
\(\mathcal R_1\) uses \(0.008\). The calculation is given in
Appendix~\ref{sec:app_rms_targeting}. The scale-aware analysis evaluates
\(\mathcal R_5\), \(\mathcal R_4\), \(\mathcal R_3\), and the full backbone
using Section~\ref{sec:scale_aware_updates}. The MRI-to-MRI analysis evaluates
\(B\), \(\mathcal R_4\), and \(\mathcal R_1\) with the same learning rates.

The approximate-RMS and scale-aware configurations use adaptation seeds
12345, 23456, and 34567; all other experiments use seed 12345. Seed-level
results are reported in Appendix~\ref{sec:app_seed_results}. Experiments run on an NVIDIA Quadro
RTX 6000 with 24\,GB memory.

The shared-LR full-backbone result follows the standard sequential training
path; restricted-region configurations use the controlled trainer. We
therefore treat the full-backbone result as a boundary case rather than an
optimizer-matched comparison.

\subsection{Evaluation Metrics}
\label{sec:evaluation_metrics}

We report the Dice similarity coefficient (DSC). For multi-class tasks,
task-level DSC is the unweighted average across evaluable foreground classes;
for ECPC-IDS, it is the DSC of the single tumor class. Let \(d_i^{(i)}\)
denote the task-level DSC on task \(i\) immediately after learning it, and let
\(d_i^{(j)}\) denote its DSC after learning task \(j\). Forgetting is
\[
F_{i\rightarrow j}
=
d_i^{(i)}-d_i^{(j)},
\qquad j>i.
\]
Positive, near-zero, and negative values indicate forgetting, retention, and
improvement, respectively. Per-structure means and standard deviations
summarize evaluation cases, not training seeds. Held-out-case bootstrap
intervals are reported in Appendix~\ref{sec:app_bootstrap}.
%===================================================================================%
\begin{table*}[t]
\floatconts
  {tab:progressive_fixed_lr}
  {\caption{Progressive UMD \(\rightarrow\) ECPC-IDS adaptation under a
  shared initial learning rate of \(0.01\). Parameter counts and displacement
  measures include trainable shared-backbone parameters only. Restricted-region
  configurations use the controlled adaptation trainer; \(\mathcal R_0\) uses
  the standard sequential training path and is treated as a boundary case.}}
  {%
    \small
    \setlength{\tabcolsep}{3pt}
    \renewcommand{\arraystretch}{1.06}
    \begin{tabular*}{\textwidth}{@{\extracolsep{\fill}}lccccc}
    \toprule
    \textbf{Trainable region} &
    \textbf{Backbone params} &
    \textbf{UMD forgetting \(\downarrow\)} &
    \textbf{ECPC-IDS DSC \(\uparrow\)} &
    \textbf{RMS \(\Delta\theta\)} &
    \textbf{Relative \(\Delta\theta\)} \\
    \midrule
    Head only
      & 0 & 0.0000 & 0.0000 & 0 & 0\% \\
    \(B\)
      & 5.53M & \(-0.0008\) & 0.2797 & 0.00442 & 34.7\% \\
    \(\mathcal R_5\)
      & 19.77M & 0.0179 & 0.6060 & 0.00433 & 34.4\% \\
    \(\mathcal R_4\)
      & 33.86M & 0.2714 & 0.7329 & 0.00486 & 37.1\% \\
    \(\mathcal R_3\)
      & 42.48M & 0.6319 & 0.7860 & 0.00554 & 39.6\% \\
    \(\mathcal R_2\)
      & 44.16M & 0.6352 & 0.8151 & 0.00545 & 36.7\% \\
    \(\mathcal R_1\)
      & 44.36M & 0.6383 & 0.8381 & 0.00551 & 35.8\% \\
    \(\mathcal R_0\) (full)
      & 44.41M & 0.6404 & 0.8658 & 0.00558 & 35.5\% \\
    \bottomrule
    \end{tabular*}
  }
\end{table*}

\section{Results and Discussion}
\label{sec:results}

Independent nnU-Net and joint training provide non-sequential
reference points, while the continual-learning baselines in
Appendix~\ref{sec:app_cl_baselines} illustrate the stability--plasticity
trade-off in the evaluated task sequence. Sequential fine-tuning
forgets previous tasks, while rehearsal improves retention by storing and
reusing earlier training examples. In clinical settings, retaining such data
may be limited by privacy, security, storage, or institutional data-governance
requirements. Stronger regularization can preserve earlier tasks but may
restrict learning of later tasks. We therefore focus the main analysis on
how backbone adaptation produces this trade-off.
%===========================================================================
\subsection{Expanding the Trainable Region Increases Learning and Forgetting}
\label{sec:depth_results}

Table~\ref{tab:progressive_fixed_lr} reports UMD
\(\rightarrow\) ECPC-IDS adaptation as the trainable backbone region expands
from the bottleneck toward the input and output. All trainable blocks use the
same initial learning rate of \(0.01\).

Bottleneck-only adaptation preserves UMD performance but achieves negligible
ECPC-IDS DSC. Expanding to \(\mathcal R_5\) provides only a small improvement.
A sharp transition occurs at \(\mathcal R_4\): ECPC-IDS DSC increases from
\(0.606\) to \(0.733\), while UMD forgetting increases from \(0.018\) to
\(0.271\). At \(\mathcal R_3\), forgetting increases further to \(0.632\).
Beyond this region, forgetting remains near \(0.64\), while current-task
performance continues to improve. The full-backbone result follows the
standard sequential training path and is included as a boundary case.

The change in forgetting is substantially larger than the change in average
parameter movement. Across the trainable configurations, RMS displacement
ranges from \(0.00433\) to \(0.00558\), whereas forgetting ranges from
approximately zero to \(0.640\). Relative displacement similarly remains
between \(34.4\%\) and \(39.6\%\). These results indicate that average parameter movement alone is insufficient
to describe forgetting. However, displacement is measured after training and
is not deliberately matched in these shared-learning-rate runs. We therefore
perform an approximate final-RMS comparison in the next section. Additional
task-wise heatmaps, class-wise results, and qualitative examples are provided
in Appendix~\ref{sec:app_depth_visuals}.

%===========================================================================
\subsection{Similar RMS Displacement Does Not Produce Similar Retention}
\label{sec:rms_control_results}

\begin{table*}[t]
\floatconts
  {tab:rms_control}
  {\caption{Approximate final-RMS comparison for UMD
  \(\rightarrow\) ECPC-IDS using adaptation seed 12345. The bottleneck result
  defines a \(\pm10\%\) target range. Matching is assessed at the final
  checkpoint and is not enforced during training. Displacement measures
  exclude task-specific heads.}}
  {%
    \small
    \setlength{\tabcolsep}{4pt}
    \renewcommand{\arraystretch}{1.08}
    \begin{tabular*}{\textwidth}{@{\extracolsep{\fill}}lccccc}
    \toprule
    \textbf{Trainable region} &
    \textbf{Initial LR} &
    \(\boldsymbol{D_{\mathrm{RMS}}}\) &
    \(\boldsymbol{\|\Delta\theta\|_2}\) &
    \textbf{UMD forgetting \(\downarrow\)} &
    \textbf{ECPC-IDS DSC \(\uparrow\)} \\
    \midrule
    \(B\)
      & 0.010 & 0.004424 & 10.40 & \(-0.0008\) & 0.2797 \\
    \(\mathcal R_4\)
      & 0.010 & 0.004859 & 28.28 & 0.2714 & 0.7329 \\
    \(\mathcal R_1\)
      & 0.008 & 0.004647 & 30.95 & 0.6384 & 0.8342 \\
    \bottomrule
    \end{tabular*}
  }
\end{table*}

Table~\ref{tab:rms_control} compares \(B\), \(\mathcal R_4\), and
\(\mathcal R_1\) at approximately comparable final RMS displacement.
Configurations \(B\) and \(\mathcal R_4\) use initial LR \(0.01\), while
\(\mathcal R_1\) uses \(0.008\) to approach the bottleneck target. The
learning-rate calculation is provided in
Appendix~\ref{sec:app_rms_targeting}. Although final RMS displacement ranges only from \(0.00442\) to \(0.00486\),
forgetting increases from approximately zero for \(B\) to \(0.271\) for
\(\mathcal R_4\) and \(0.638\) for \(\mathcal R_1\). Over the same
configurations, ECPC-IDS DSC increases from \(0.280\) to \(0.733\) and
\(0.834\), respectively.

Configuration \(\mathcal R_1\) can also be compared with its LR \(0.01\)
counterpart in Table~\ref{tab:progressive_fixed_lr}. Reducing the initial LR
to \(0.008\) lowers RMS displacement from \(0.00551\) to \(0.00465\), while
forgetting remains nearly unchanged at \(0.6383\) and \(0.6384\),
respectively.

Thus, comparable average per-parameter movement does not produce comparable
retention across adaptation extents. This control does not match total
\(\ell_2\) displacement or movement within individual blocks. The next
experiment examines how block-specific learning-rate calibration changes the
magnitude and depth-wise distribution of these updates.

%===========================================================================

\begin{figure}[t]
\floatconts
  {fig:scale_control}
  {\caption{Shared and scale-aware learning rates across adaptation extents.
(a) UMD forgetting and (b) ECPC-IDS current-task DSC for
\(\mathcal R_5\), \(\mathcal R_4\), \(\mathcal R_3\), and
\(\mathcal R_0\), where \(\mathcal R_0\) denotes the full backbone.}}
  {%
    \includegraphics[width=\linewidth]{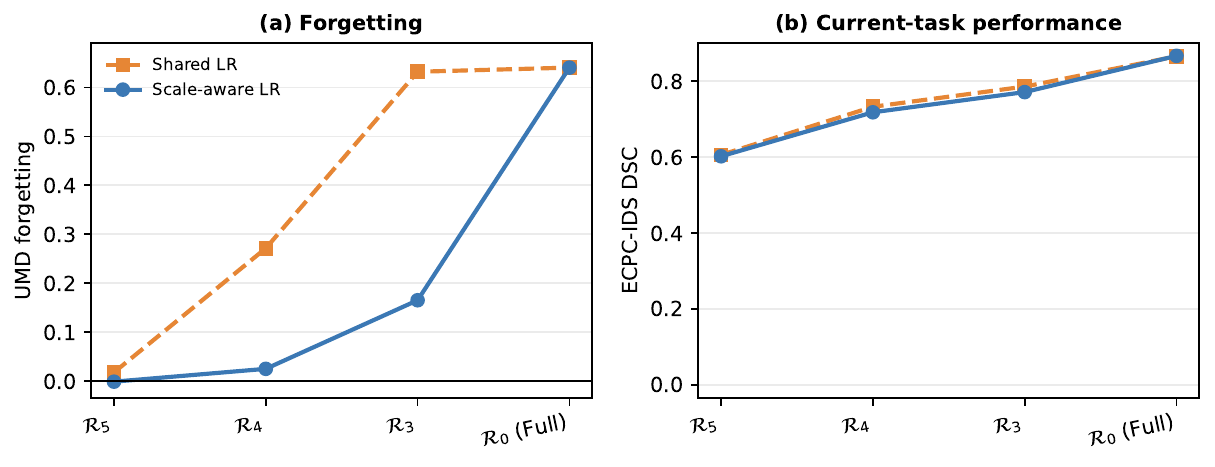}
  }
\end{figure}

\subsection{Block-Specific Learning Rates Change the Trade-off}
\label{sec:scale_aware_results}

Figure~\ref{fig:scale_control} and
Table~\ref{tab:scale_aware_control} compare shared and scale-aware learning
rates around the sharp forgetting transition. Within each restricted-region
pair, the trainable blocks and controlled training procedure are identical;
only the allocation of learning rates differs. The full-backbone results are
included as boundary cases but use different training paths.

\begin{table*}[t]
\floatconts
  {tab:scale_aware_control}
  {\caption{Shared and scale-aware learning rates for UMD $\rightarrow$
  ECPC-IDS. Values are mean $\pm$ sample standard deviation over three
  adaptation seeds and describe training-seed variability, not held-out-case
  variability. Scale-aware rates are calibrated separately for each seed from
  20 gradient-only batches. Within each region, both allocations use the same
  trainable blocks and the same controlled training procedure. Bold marks the
  lower forgetting where the two allocations differ substantially. The
  $\mathcal{R}_5$ shared-LR DSC reflects a correction to the evaluation
  preprocessing (PET/CT settings replacing the originally applied MRI
  settings); training was unaffected.}}
  {%
    \small
    \setlength{\tabcolsep}{3.5pt}
    \renewcommand{\arraystretch}{1.06}
    \begin{tabular*}{\textwidth}{@{\extracolsep{\fill}}llcc}
    \toprule
    \textbf{Region} &
    \textbf{LR allocation} &
    \textbf{UMD forgetting ($\downarrow$)} &
    \textbf{ECPC-IDS DSC ($\uparrow$)} \\
    \midrule
    $\mathcal{R}_5$
      & Shared      & $0.020 \pm 0.003$ & $0.602 \pm 0.003$ \\
    $\mathcal{R}_5$
      & Scale-aware & $0.003 \pm 0.004$ & $0.600 \pm 0.008$ \\
    \addlinespace
    $\mathcal{R}_4$
      & Shared      & $0.363 \pm 0.079$ & $0.740 \pm 0.010$ \\
    $\mathcal{R}_4$
      & Scale-aware & $\mathbf{0.045 \pm 0.018}$ & $0.724 \pm 0.005$ \\
    \addlinespace
    $\mathcal{R}_3$
      & Shared      & $0.624 \pm 0.008$ & $0.785 \pm 0.005$ \\
    $\mathcal{R}_3$
      & Scale-aware & $\mathbf{0.199 \pm 0.035}$ & $0.773 \pm 0.002$ \\
    \addlinespace
    $\mathcal{R}_0$ (full)
      & Shared      & $0.640 \pm 0.000$ & $0.865 \pm 0.004$ \\
    $\mathcal{R}_0$ (full)
      & Scale-aware & $0.640 \pm 0.000$ & $0.863 \pm 0.004$ \\
    \bottomrule
    \end{tabular*}
  }
\end{table*}

Block-specific learning rates substantially improve the trade-off in the
restricted regions. For \(\mathcal R_5\), ECPC-IDS DSC is unchanged
(\(0.602\) vs.\ \(0.600\)) while forgetting falls from \(0.020\pm0.003\)
to \(0.003\pm0.004\) without additional forgetting. For \(\mathcal R_4\)
and \(\mathcal R_3\), forgetting decreases from \(0.363\pm0.079\) to
\(0.045\pm0.018\) and from \(0.624\pm0.008\) to \(0.199\pm0.035\),
respectively, with only small reductions in ECPC-IDS DSC (\(0.740\) to
\(0.724\) and \(0.785\) to \(0.773\)). Notably, the improvement at
\(\mathcal R_5\) occurs despite a lower \(D_{\mathrm{RMS}}\), showing that
smaller average parameter movement does not necessarily imply weaker
current-task learning. Across the restricted regions, calibration therefore
produces a more favorable balance between retaining UMD and learning ECPC-IDS
than the shared learning rate.

For \(\mathcal R_4\), calibration shifts parameter movement toward the
bottleneck. The squared-displacement share of \(E_4+D_4\) decreases from
\(55.9\%\) to \(12.2\%\), while the bottleneck share increases from
\(12.2\%\) to \(55.1\%\). Nevertheless, forgetting under scale-aware rates
still increases as adaptation expands, from approximately zero at
\(\mathcal R_5\) to \(0.025\) at \(\mathcal R_4\) and \(0.165\) at
\(\mathcal R_3\) (seed 12345). For the full backbone, scale-aware rates reduce
RMS displacement from \(0.00558\) to \(0.00222\), but forgetting remains
\(0.640\). Because the full-backbone configurations use different training
paths, this comparison is descriptive rather than optimizer-matched. Overall,
calibration changes both update magnitude and its depth-wise distribution. It
therefore demonstrates that block-specific scaling can alter the
stability--plasticity trade-off, but does not isolate the effect of update
allocation alone. Complete block-level results and learning-rate multipliers
are provided in Appendix~\ref{sec:app_displacement}.

%===========================================================================
\subsection{Allocation, Not Just Smaller Updates}
\label{sec:uniform_lr_control}

Because scale-aware calibration simultaneously reduces overall update
magnitude and redistributes it across blocks, it is unclear how much of
the retention gain comes from the reallocation itself versus simply using a
smaller effective learning rate. To isolate this, we compare scale-aware
rates against uniformly reduced shared learning rates at \(\mathcal{R}_4\)
(Table~\ref{tab:uniform_lr_control}, seed 12345).

\begin{table}[t]
\floatconts
  {tab:uniform_lr_control}
  {\caption{Shared, uniformly lowered, and scale-aware learning rates at
  \(\mathcal{R}_4\) (UMD \(\rightarrow\) ECPC-IDS, seed 12345).
  \(q_{E_4+D_4}\) and \(q_B\) are the shares of total squared backbone
  displacement in the outermost trainable pair and the bottleneck,
  respectively.}}
  {%
    \small
    \setlength{\tabcolsep}{3pt}
    \renewcommand{\arraystretch}{1.06}
    \begin{tabular*}{\columnwidth}{@{\extracolsep{\fill}}lcccc}
    \toprule
    \textbf{LR setting} &
    \(\boldsymbol{D_{\mathrm{RMS}}}\) &
    \(\boldsymbol{q_{E_4+D_4}}\) &
    \textbf{Forget.\(\downarrow\)} &
    \textbf{DSC\(\uparrow\)} \\
    \midrule
    Shared 0.010   & 0.00486 & 55.9\% & 0.271 & 0.733 \\
    Uniform 0.005  & 0.00304 & 65.7\% & 0.140 & 0.731 \\
    Uniform 0.0025 & 0.00199 & 76.4\% & 0.098 & 0.738 \\
    Scale-aware    & 0.00227 & 12.2\% & \textbf{0.025} & 0.718 \\
    \bottomrule
    \end{tabular*}
  }
\end{table}

Lowering the shared learning rate does reduce forgetting (\(0.271
\rightarrow 0.098\) at LR \(0.0025\)), but it \emph{concentrates} even more
movement in the outermost pair: the \(E_4{+}D_4\) share rises from
\(55.9\%\) to \(76.4\%\) as the learning rate decreases. Scale-aware rates
instead redirect movement toward the bottleneck (\(q_B\) rises from
\(12.2\%\) to \(55.1\%\)), and forgetting falls to \(0.025\)---roughly four
times lower than the best uniform rate---even though \(D_{\mathrm{RMS}}\)
is slightly \emph{higher} than at LR \(0.0025\). This comes at a cost of
about two Dice points (\(0.718\) vs.\ \(0.738\)). The reduction in
forgetting therefore cannot be explained by smaller aggregate updates alone.
Over three seeds, scale-aware forgetting at \(\mathcal R_4\) is
\(0.045\pm0.018\), which remains well below \(0.098\) and confirms this
conclusion. Retention depends on how updates are distributed across blocks,
not only on their total magnitude.

%===========================================================================
\subsection{MRI-to-MRI Analysis}
\label{sec:cross_transition_results}

We next examine whether the result is limited to the large MRI-to-PET/CT shift
between UMD and ECPC-IDS. Table~\ref{tab:endo_replication} reports the same
representative adaptation regions for UMD
\(\rightarrow\) UT-EndoMRI, where both tasks use MRI. \(B\) and \(\mathcal{R}_4\) have nearly identical final RMS parameter
displacement (\(0.00681\) and \(0.00673\)), but produce substantially
different UMD forgetting (\(0.059\) and \(0.638\), respectively).
Compared with \(B\), \(\mathcal{R}_4\) increases UT-EndoMRI DSC from
\(0.099\) to \(0.243\), while \(\mathcal{R}_1\) increases it further to
\(0.302\) but produces similarly severe UMD forgetting. This exploratory MRI-to-MRI analysis shows the same association between
broader adaptation and greater forgetting. UT-EndoMRI is also challenging
when trained independently, with the single-task nnU-Net obtaining low DSC
for several structures, particularly the ovaries and cysts
(Appendix~\ref{sec:app_cl_baselines}). Its comparatively low current-task DSC
should therefore not be attributed solely to continual adaptation. Because
UMD remains the first task and T1FS selection included the evaluation results,
this analysis is treated as supporting evidence rather than an independent
confirmation.

\begin{table}[t]
\floatconts
  {tab:endo_replication}
  {\caption{Secondary UMD \(\rightarrow\) UT-EndoMRI analysis. Learning-rate
settings are transferred from the primary analysis without retuning. Sequence
selection included exploratory inspection of the held-out evaluation cohort.}}
  {%
    
    \resizebox{\linewidth}{!}{%
    \begin{tabular}{lcccc}
    \toprule
    \textbf{Region} &
    \textbf{LR} &
    \textbf{RMS} &
    \textbf{\begin{tabular}[c]{@{}c@{}}UMD Forgetting\\ \(\downarrow\)\end{tabular}} &
    \textbf{\begin{tabular}[c]{@{}c@{}}UT-EndoMRI DSC\\ \(\uparrow\)\end{tabular}} \\
    \midrule
    \(B\)
      & 0.010 & 0.00681 & 0.0592 & 0.0985 \\
    \(\mathcal R_4\)
      & 0.010 & 0.00673 & 0.6380 & 0.2435 \\
    \(\mathcal R_1\)
      & 0.008 & 0.00573 & 0.6362 & 0.3019 \\
    \bottomrule
    \end{tabular}%
    }%
  }
\end{table}

%===========================================================================
\section{Conclusion}
\label{sec:conclusion}

We presented a controlled framework for revealing how adaptation across
encoder--decoder depth shapes catastrophic forgetting. By combining nested
trainable regions, parameter-displacement analysis, and gradient-calibrated
learning rates, we show that similar average parameter movement can produce
markedly different retention depending on where updates occur. Block-specific
scaling substantially improves retention at intermediate adaptation extents,
demonstrating that both the location and scale of adaptation are central to
the stability--plasticity trade-off. This is especially important in gynecological imaging, where segmentation
tasks vary across modality, anatomy, pathology, and annotation structure,
while privacy and data-governance constraints may prevent earlier datasets
from being retained. Our findings provide a strong basis for future
continual-learning methods that control where and how strongly a model adapts.
Although evaluated with one architecture and a fixed UMD-first task order,
the repeated-seed and MRI-to-MRI analyses support the consistency of the
observed pattern.

\bibliography{ref}

\section{Additional Dataset Details}
\label{sec:app_dataset_details}

\subsection{Dataset Statistics}
\label{sec:app_dataset_stats}

\begin{figure*}[t]
\centering
\resizebox{\textwidth}{!}{%
\begin{tabular}{ccc}
\includegraphics[height=3.5cm]{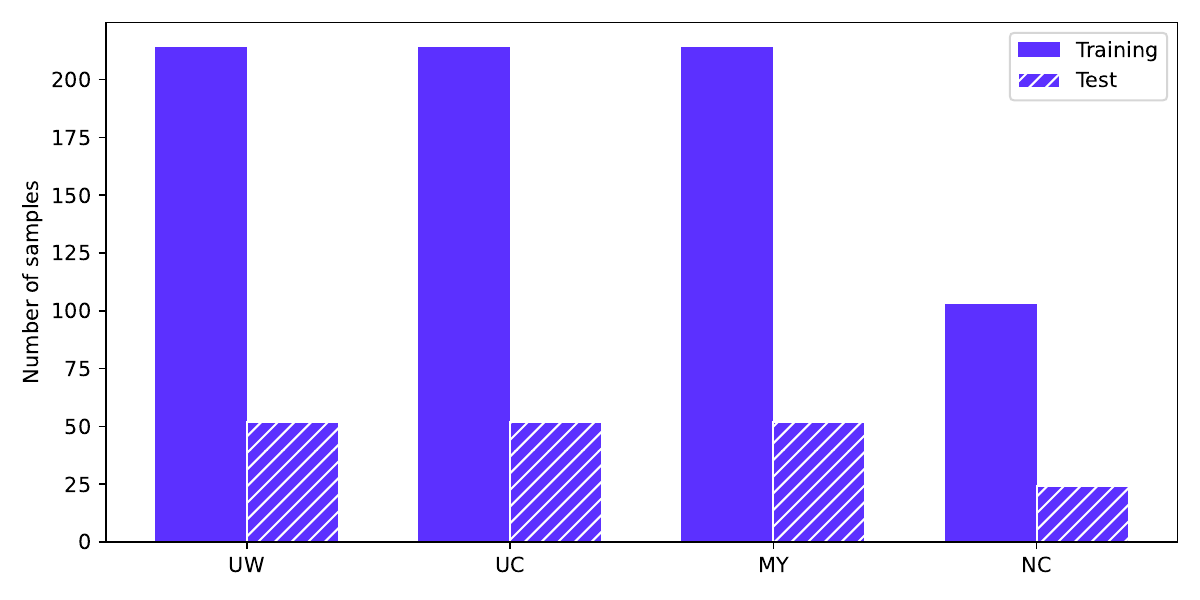} &
\includegraphics[height=3.5cm]{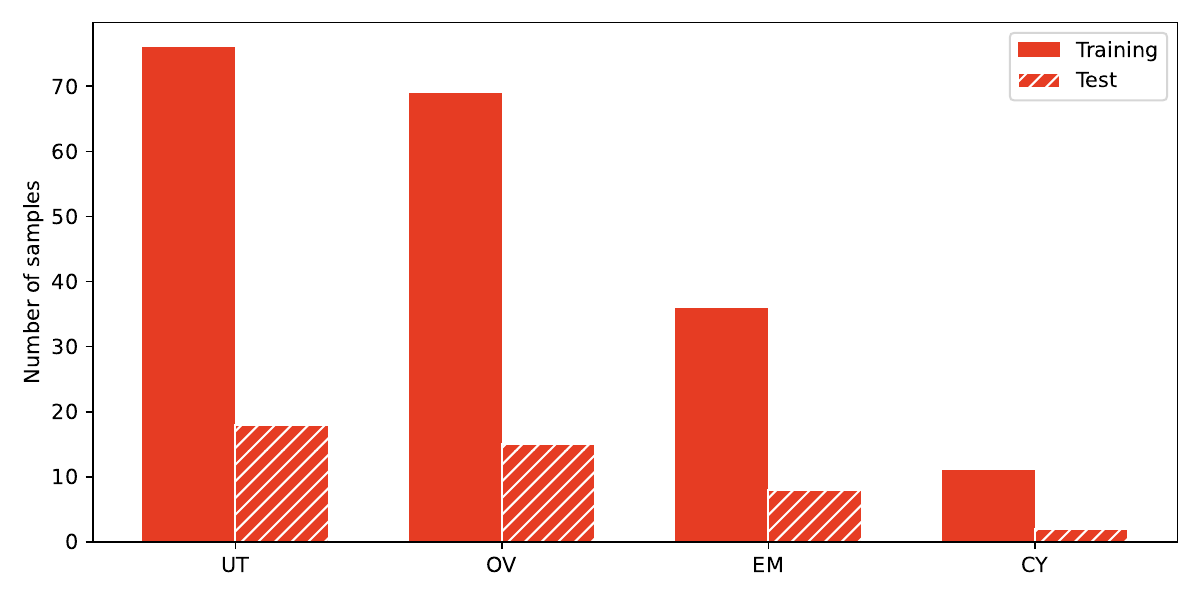} &
\includegraphics[height=3.5cm]{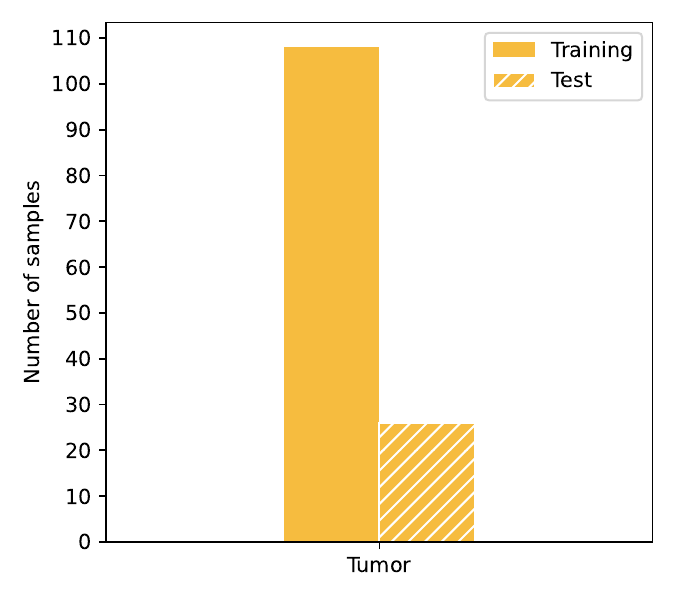} \\
(a) & (b) & (c)
\end{tabular}%
}
\caption{Dataset statistics across the 80/20 train--test splits. (a) UMD annotation distribution. (b) UT-EndoMRI sequence and label distribution. (c) ECPC-IDS preprocessing statistics.}
\label{fig:dataset_stats}
\end{figure*}

Figure~\ref{fig:dataset_stats} summarizes annotation and preprocessing statistics for the three datasets, providing additional context for their heterogeneity.

\subsection{Exploratory UT-EndoMRI Sequence Comparison}
\label{sec:app_endomri_selection}

UT-EndoMRI contains T1, T1FS, T2, and T2FS acquisitions, with sequence
availability varying across cases. Table~\ref{tab:endomri_sequence_dice}
reports the exploratory comparison used when selecting the input sequence.
The values were computed on held-out evaluation subsets whose composition
varies with sequence availability. Because these results were inspected
during sequence selection, they should not be interpreted as an independent
evaluation of the selected T1FS configuration.

T1FS also achieved the highest macro-averaged foreground DSC on the internal validation data: \(0.2983\), compared with \(0.2956\) for T1FS+T2,
\(0.2811\) for all sequences, and \(0.2160\) for T2. Because this validation
analysis was formalized after inspection of the held-out results, the sequence
comparison and subsequent MRI-to-MRI analysis are treated as exploratory.

\begin{table}[t]
\centering
\caption{Exploratory single-task results on the UT-EndoMRI held-out
evaluation subsets. Values are case-level mean DSC \(\pm\) SD. Subset
composition varies with sequence availability.}
\label{tab:endomri_sequence_dice}
\resizebox{\linewidth}{!}{%
\begin{tabular}{lcccc}
\toprule
\textbf{Structure} & \textbf{T1FS} & \textbf{T2} & \textbf{T1FS \& T2} & \textbf{All} \\
\midrule
UT & 0.6258 $\pm$ 0.2553 & 0.5443 $\pm$ 0.3392 & 0.5605 $\pm$ 0.2899 & 0.5218 $\pm$ 0.3334 \\
EM & 0.3419 $\pm$ 0.3361 & 0.2144 $\pm$ 0.3287 & 0.2353 $\pm$ 0.3426 & 0.2199 $\pm$ 0.3184 \\
OV & 0.1553 $\pm$ 0.2448 & 0.1931 $\pm$ 0.2304 & 0.0529 $\pm$ 0.0836 & 0.0606 $\pm$ 0.1463 \\
CY & 0.0273 $\pm$ 0.0864 & 0.0649 $\pm$ 0.2220 & 0.0195 $\pm$ 0.0319 & 0.0499 $\pm$ 0.1223 \\
\bottomrule
\end{tabular}%
}
\end{table}

%=================================================================%

\section{Single, Joint, and Continual-Learning Performance}
\label{sec:app_cl_baselines}

We first compare independent single-task nnU-Net models with joint MultiTalent~\cite{multitalent} training to establish reference performance when datasets are learned separately or simultaneously. As shown in Table~\ref{tab:baseline_vs_joint}, independent training achieves higher DSC for most structures, particularly on UMD and UT-EndoMRI, while performance on ECPC-IDS is similar. These results illustrate the difficulty of jointly representing the heterogeneous imaging and segmentation tasks considered in this study.

\begin{table}[t]
\centering
\caption{Segmentation performance (DSC) under independent (nnU-Net) and joint (MultiTalent) training. Values are mean $\pm$ SD. Best results per structure are shown in bold.}
\label{tab:baseline_vs_joint}
\resizebox{\columnwidth}{!}{%
\begin{tabular}{lcc}
\toprule
\textbf{Structure} & \textbf{nnU-Net (Single)} & \textbf{MultiTalent (Joint)} \\
\midrule

\multicolumn{3}{l}{\textbf{UMD}} \\
UW & \textbf{0.8339 $\pm$ 0.0920} & 0.6237 $\pm$ 0.1082 \\
UC & \textbf{0.7421 $\pm$ 0.1778} & 0.6546 $\pm$ 0.1704 \\
MY & \textbf{0.6481 $\pm$ 0.2868} & 0.3986 $\pm$ 0.3412 \\
NC & \textbf{0.3875 $\pm$ 0.3689} & 0.1618 $\pm$ 0.2393 \\
\midrule

\multicolumn{3}{l}{\textbf{ECPC-IDS}} \\
Tumor & \textbf{0.8817 $\pm$ 0.1402} & 0.8726 $\pm$ 0.1441 \\
\midrule

\multicolumn{3}{l}{\textbf{UT-EndoMRI}} \\
UT & \textbf{0.6258 $\pm$ 0.2553} & 0.4370 $\pm$ 0.2281 \\
EM & \textbf{0.3419 $\pm$ 0.3361} & 0.1647 $\pm$ 0.2766 \\
OV & \textbf{0.1553 $\pm$ 0.2448} & 0.0711 $\pm$ 0.1130 \\
CY & \textbf{0.0273 $\pm$ 0.0864} & 0.0158 $\pm$ 0.0664 \\
\bottomrule
\end{tabular}%
}
\end{table}

We next evaluate representative continual learning strategies on the fixed
three-task sequence UMD $\rightarrow$ ECPC-IDS $\rightarrow$ UT-EndoMRI.
We consider sequential fine-tuning, replay-based rehearsal, distillation-based
Learning without Forgetting (LwF), and regularization-based Elastic Weight
Consolidation (EWC). Figure~\ref{fig:cl_heatmaps_all} shows task-wise
segmentation performance across the sequence, while
Table~\ref{tab:cl_forgetting} reports forgetting on previously learned tasks.
Multiple hyperparameter settings are included to illustrate the
stability--plasticity behavior of each method.

\begin{figure*}[htbp]
\floatconts
  {fig:cl_heatmaps_all}
  {\caption{Task-wise DSC heatmaps for continual learning methods and
  hyperparameter variants. (a) Sequential fine-tuning;
  (b)--(c) rehearsal with 10\% and 25\% memory, respectively;
  (d)--(f) LwF with $T=1$, $T=2$, and $T=4$, respectively;
  (g)--(i) EWC with $\lambda=0.01$, $\lambda=0.1$, and
  $\lambda=1.0$, respectively.}}
  {%
    \setlength{\tabcolsep}{3pt}
    \renewcommand{\arraystretch}{1.0}

    \begin{tabular}{ccc}

    \includegraphics[width=0.30\textwidth]{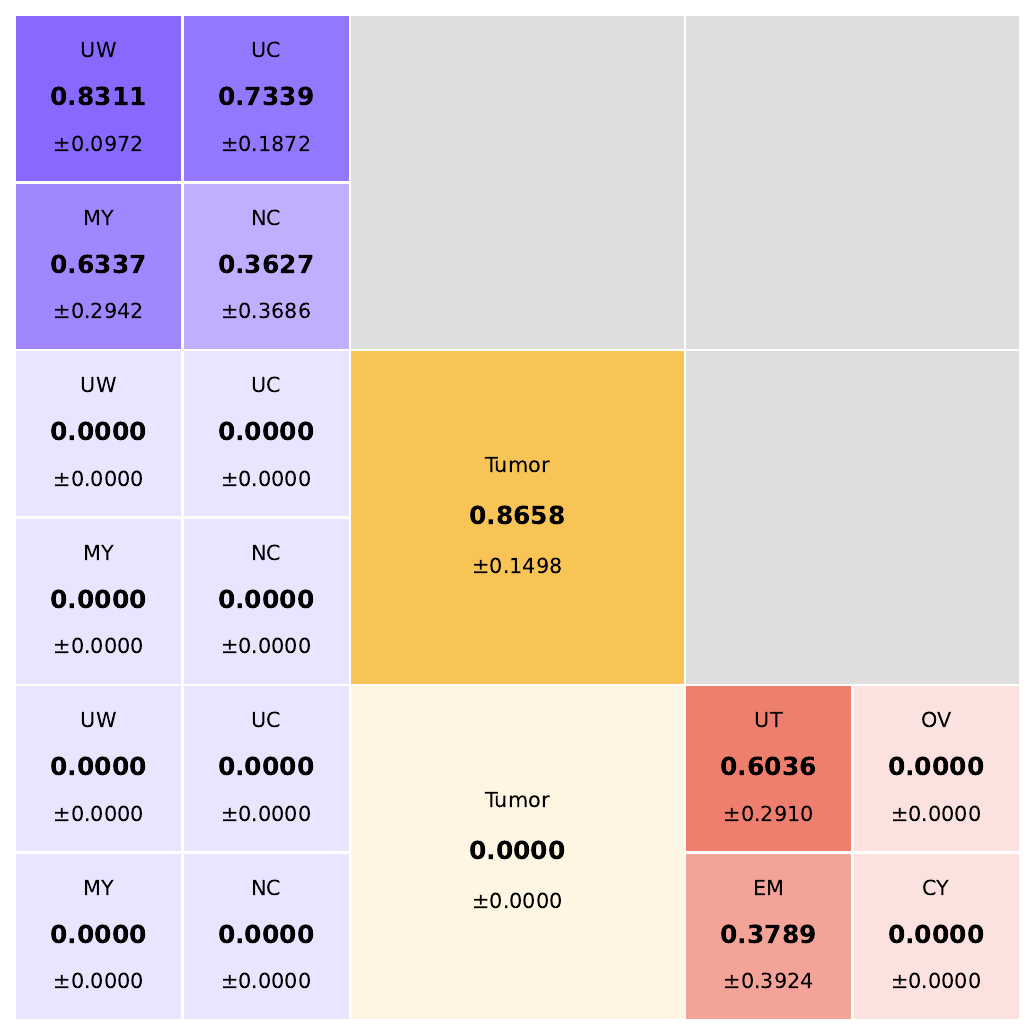} &
    \includegraphics[width=0.30\textwidth]{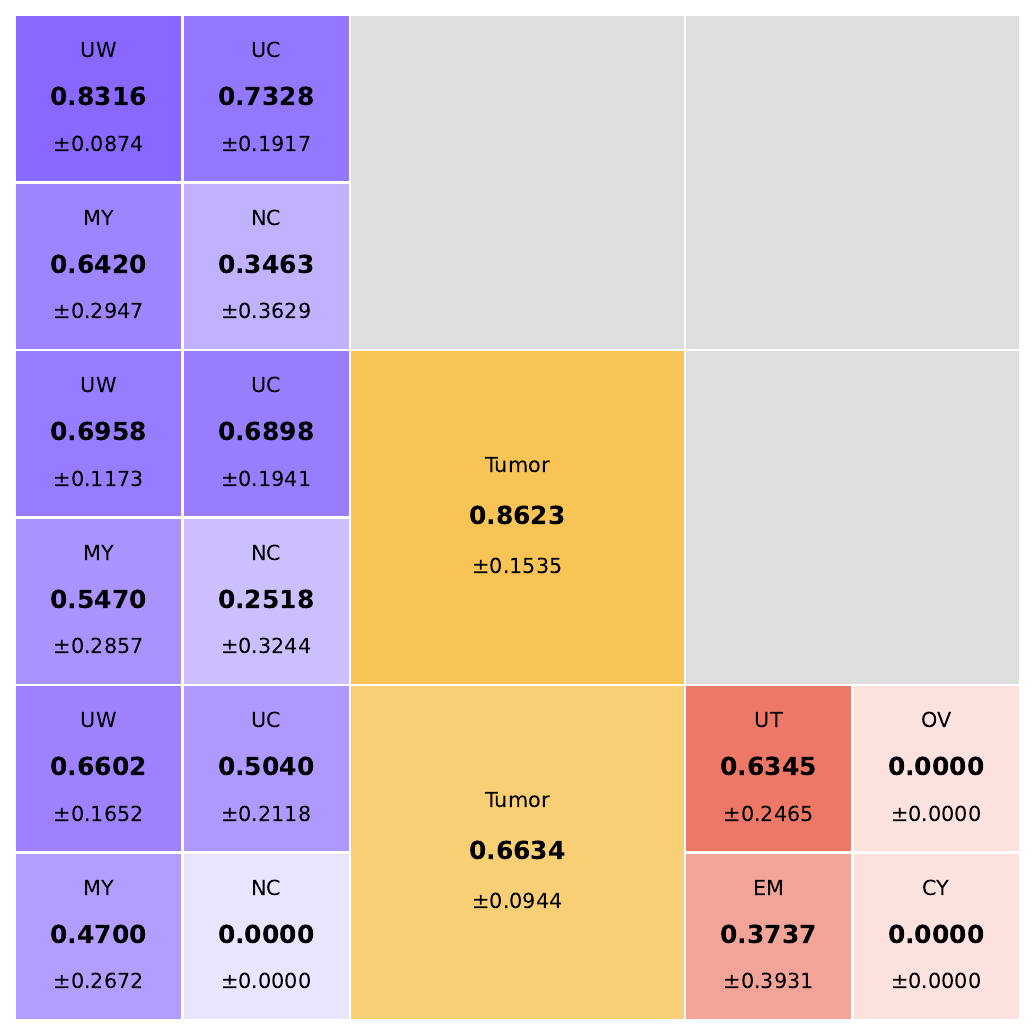} &
    \includegraphics[width=0.30\textwidth]{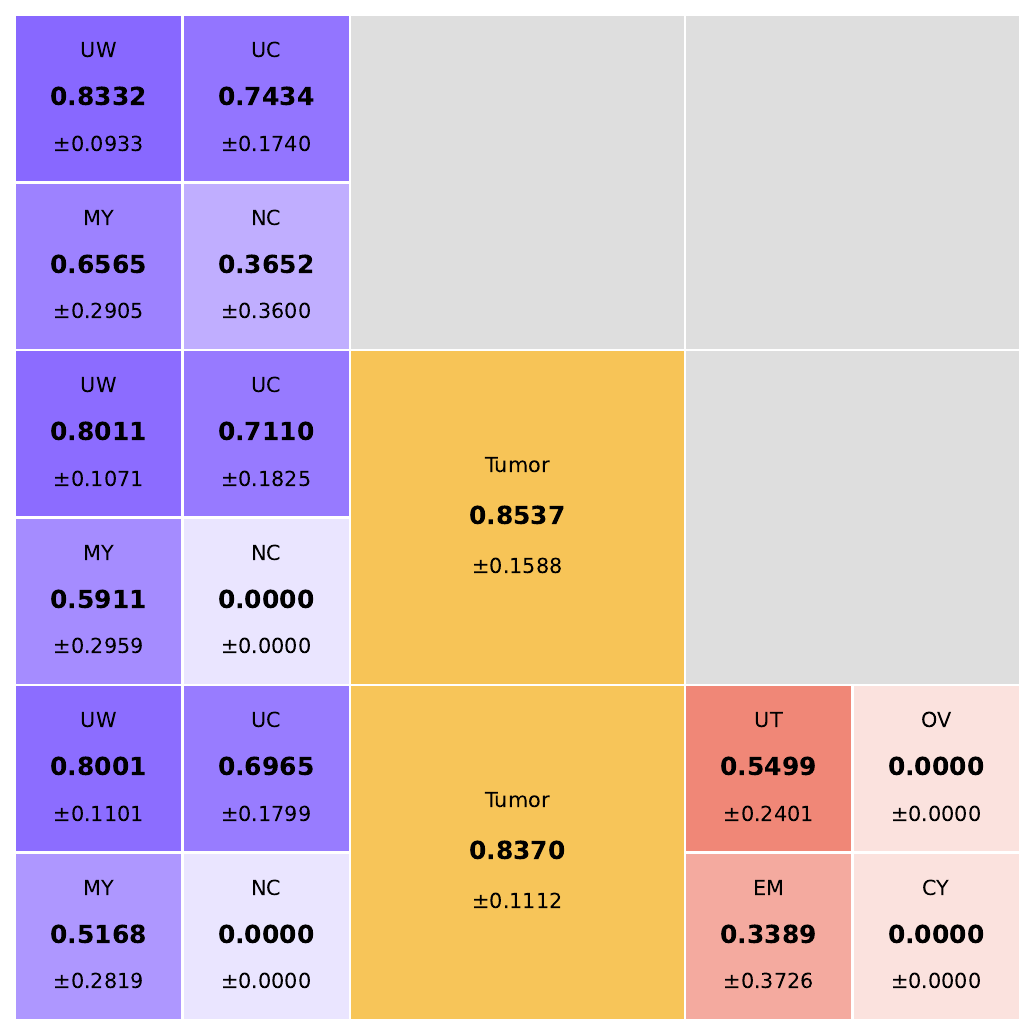} \\
    (a) Sequential &
    (b) Rehearsal 10\% &
    (c) Rehearsal 25\% \\[2mm]

    \includegraphics[width=0.30\textwidth]{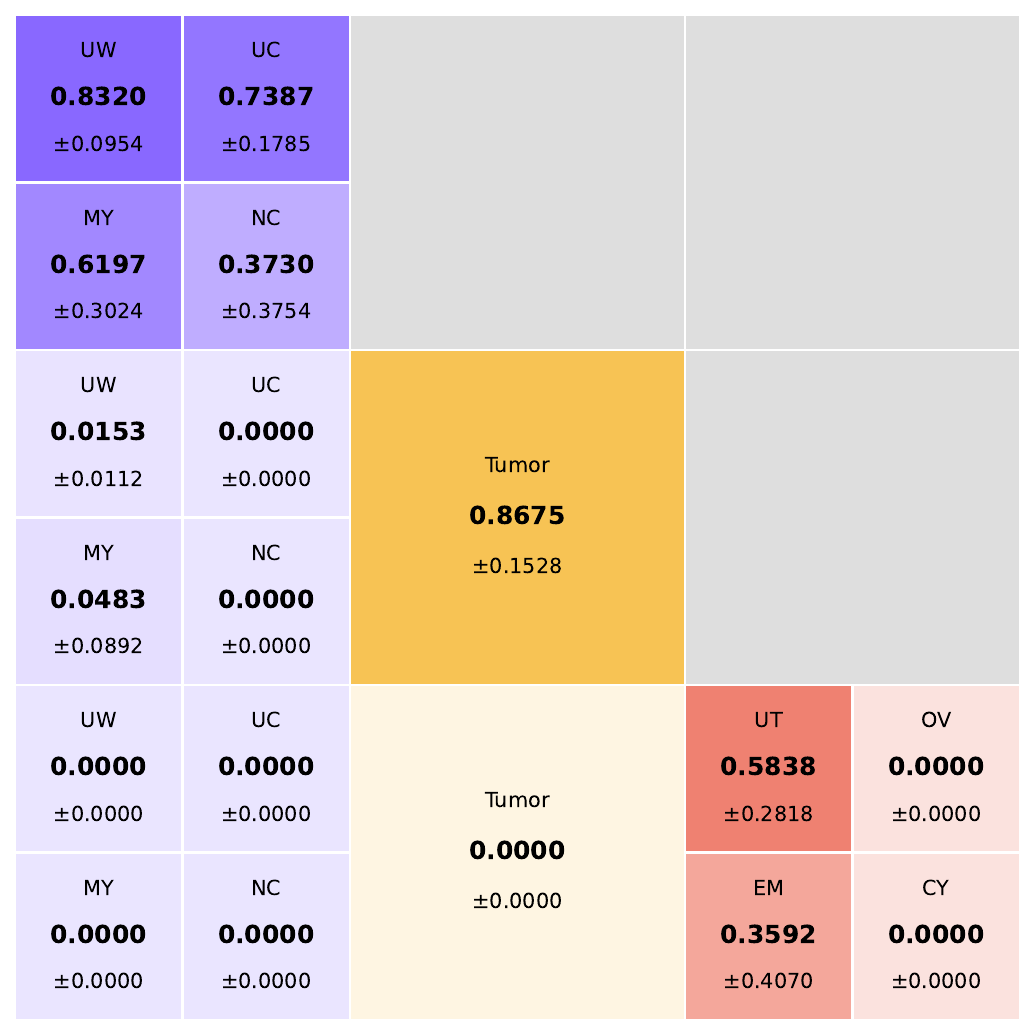} &
    \includegraphics[width=0.30\textwidth]{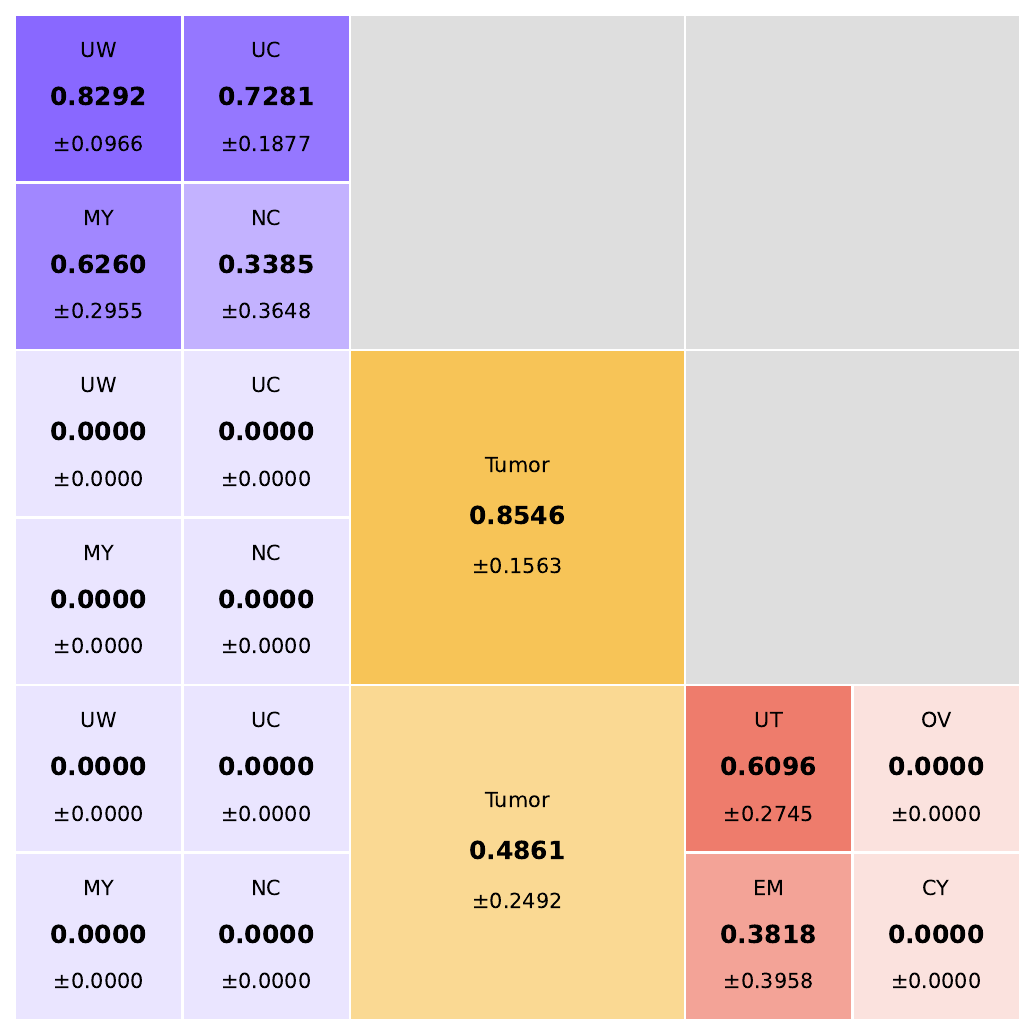} &
    \includegraphics[width=0.30\textwidth]{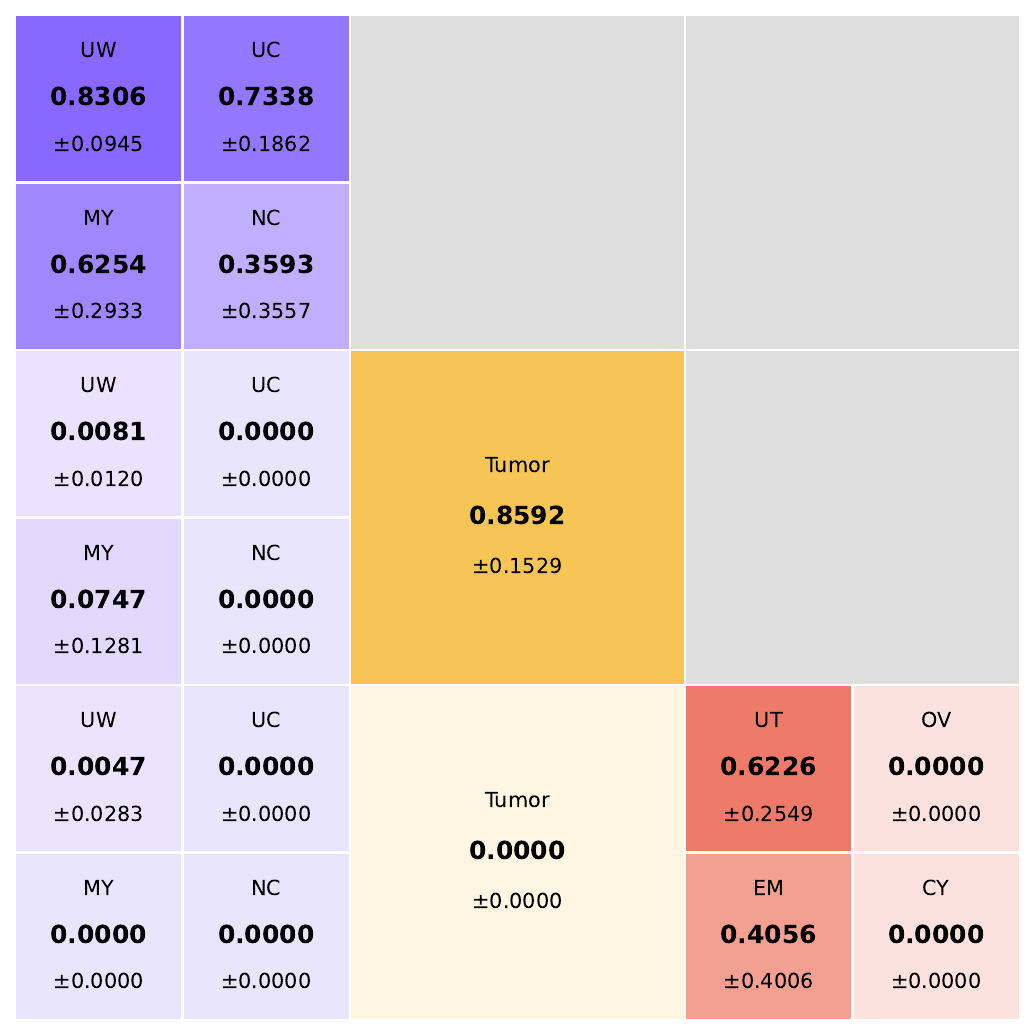} \\
    (d) LwF, $T=1$ &
    (e) LwF, $T=2$ &
    (f) LwF, $T=4$ \\[2mm]

    \includegraphics[width=0.30\textwidth]{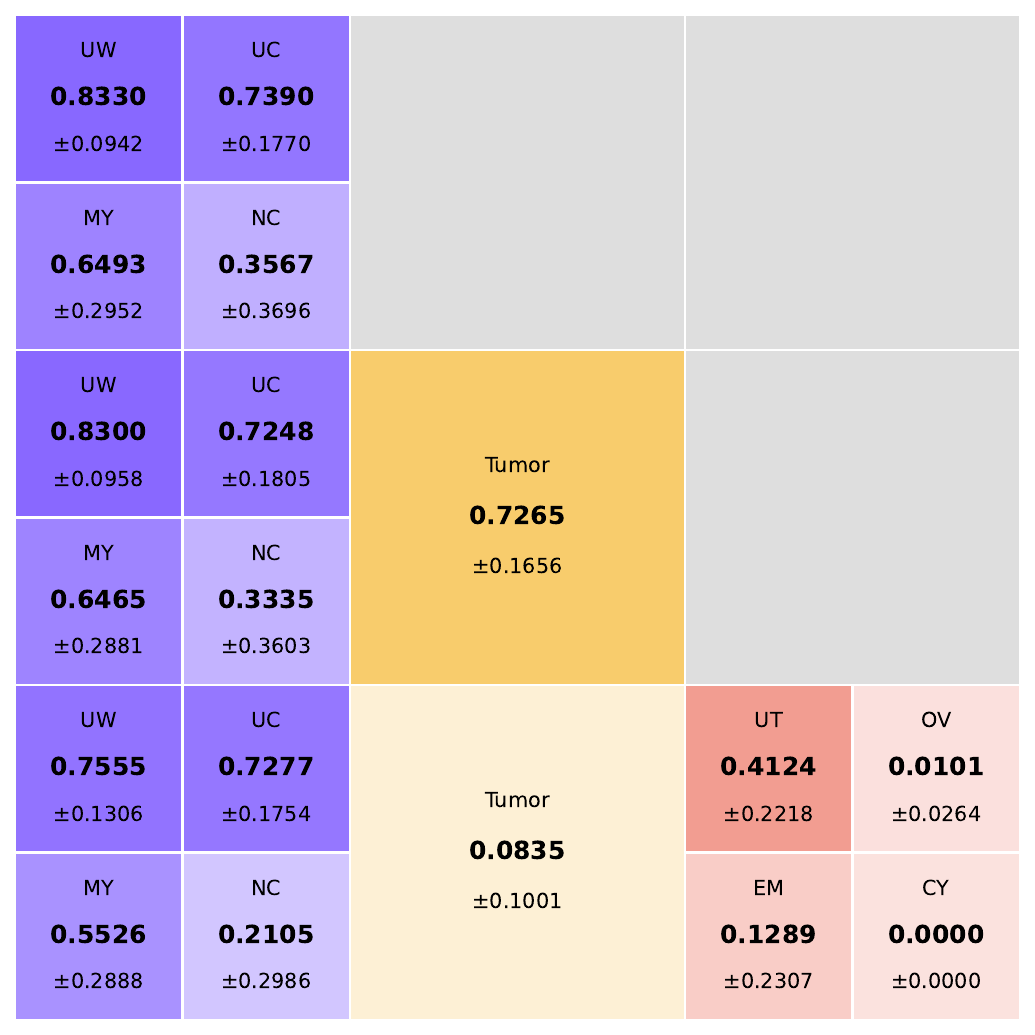} &
    \includegraphics[width=0.30\textwidth]{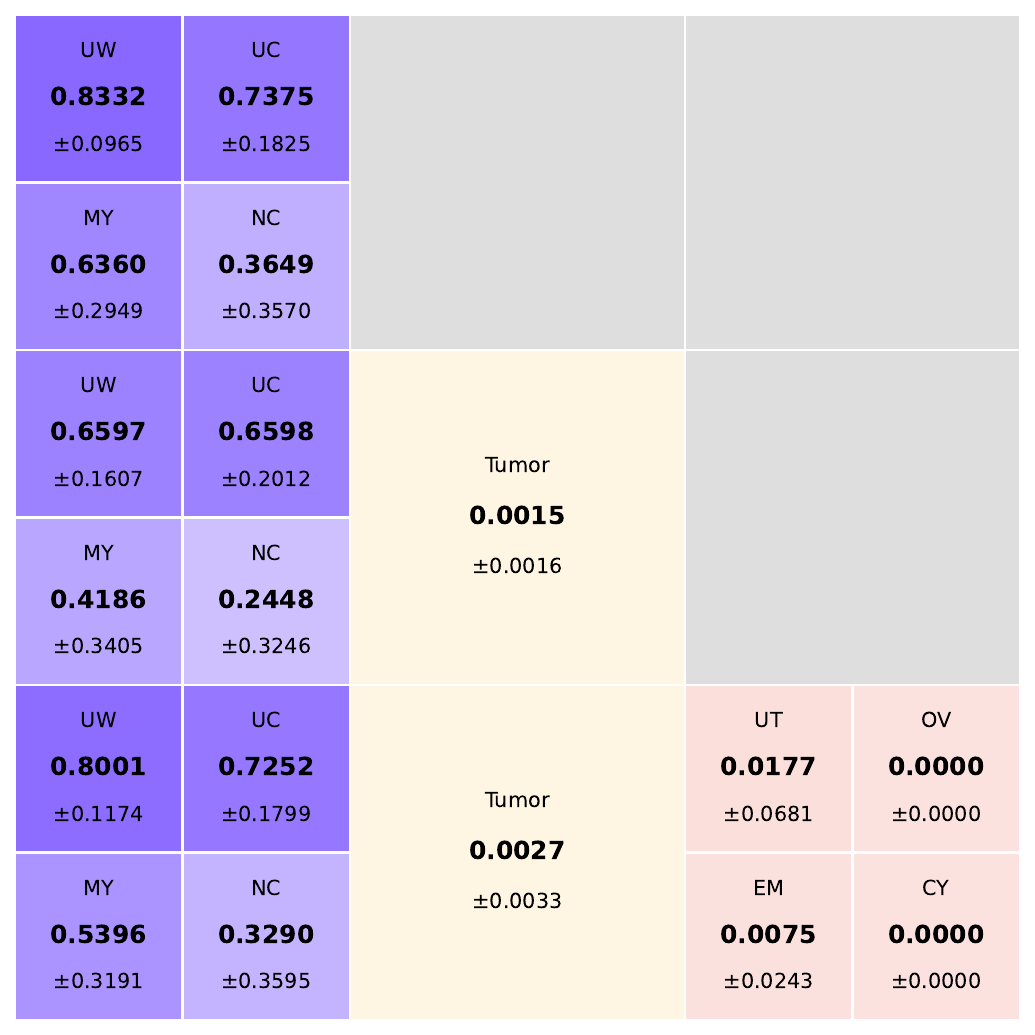} &
    \includegraphics[width=0.30\textwidth]{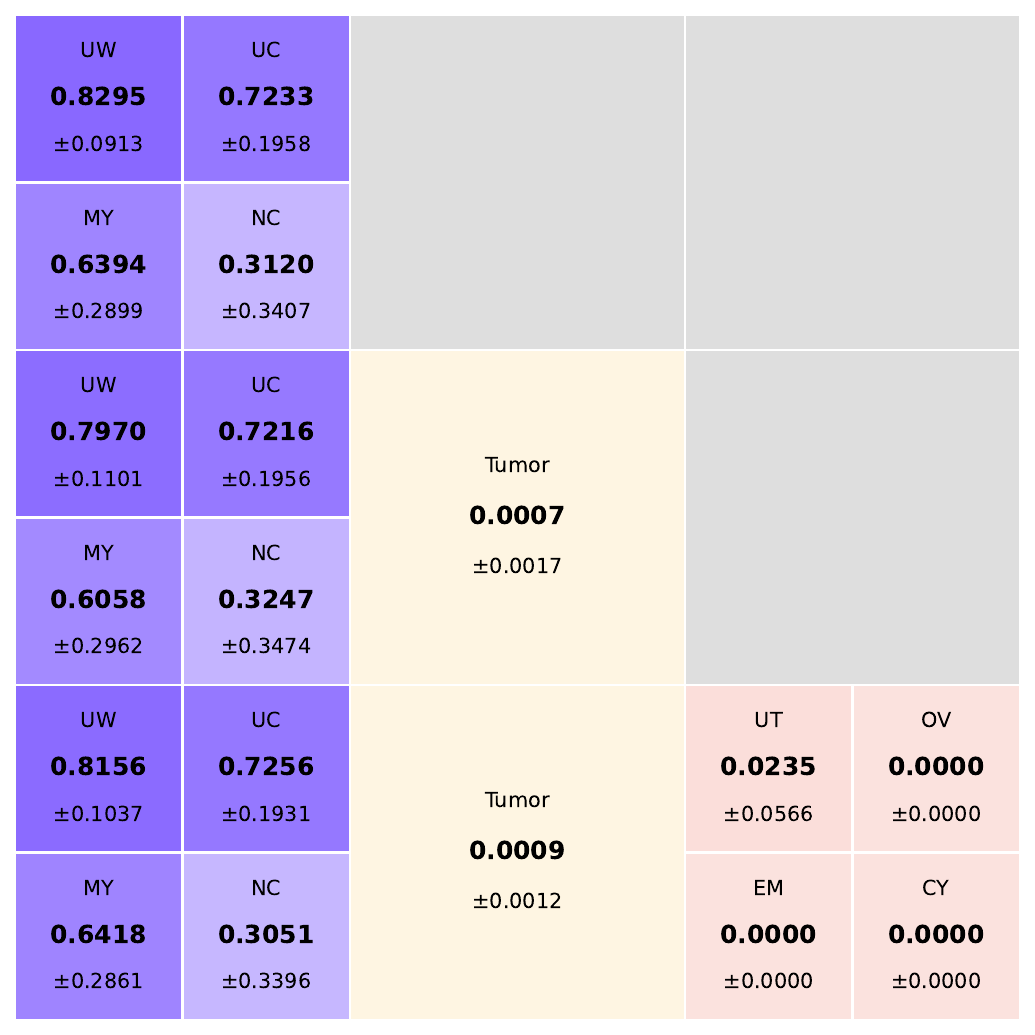} \\
    (g) EWC, $\lambda=0.01$ &
    (h) EWC, $\lambda=0.1$ &
    (i) EWC, $\lambda=1.0$

    \end{tabular}
  }
\end{figure*}

\begin{table}[htbp]
\floatconts
  {tab:cl_forgetting}
  {\caption{Average forgetting across continual learning methods and
  hyperparameter variants. Lower values indicate better retention.}}
  {%
    \small
    \setlength{\tabcolsep}{3pt}
    \renewcommand{\arraystretch}{1.08}

    \begin{tabular*}{\columnwidth}
      {@{\extracolsep{\fill}}lccc}
    \toprule
    \textbf{Method} &
    \textbf{UMD} &
    \textbf{ECPC-IDS} &
    \textbf{Average} \\
    \midrule

    Sequential Fine-tuning
    & 0.6404
    & 0.8658
    & 0.7531 \\

    \addlinespace

    Rehearsal (10\%)
    & 0.2296
    & 0.1989
    & 0.2143 \\

    Rehearsal (25\%)
    & 0.1462
    & 0.0167
    & 0.0815 \\

    \addlinespace

    LwF ($T=1$)
    & 0.6408
    & 0.8675
    & 0.7541 \\

    LwF ($T=2$)
    & 0.6305
    & 0.3685
    & 0.4995 \\

    LwF ($T=4$)
    & 0.6361
    & 0.8592
    & 0.7476 \\

    \addlinespace

    EWC ($\lambda=0.01$)
    & 0.0830
    & 0.6430
    & 0.3630 \\

    EWC ($\lambda=0.1$)
    & 0.0445
    & -0.0011
    & 0.0217 \\

    EWC ($\lambda=1.0$)
    & 0.0040
    & -0.0002
    & 0.0019 \\

    \bottomrule
    \end{tabular*}
  }
\end{table}

Forgetting alone does not fully characterize continual-learning performance, as low forgetting may result from limited adaptation to the new task. Sequential fine-tuning shows severe forgetting, whereas rehearsal improves retention while maintaining stronger adaptation to subsequent tasks. Increasing the EWC regularization strength substantially reduces measured
forgetting, but the corresponding heatmaps show that this stability is obtained at the cost of reduced current-task learning. LwF provides only partial stabilization across the evaluated temperature settings. These results illustrate the stability--plasticity trade-off present in this heterogeneous sequence and motivate the depth-wise analysis in the main paper.

%===================================================%

\section{Progressive Adaptation Visualizations}
\label{sec:app_depth_visuals}

\begin{figure*}[t]
\centering
\setlength{\tabcolsep}{4pt}

\begin{tabular}{cccc}
\includegraphics[width=0.23\textwidth]{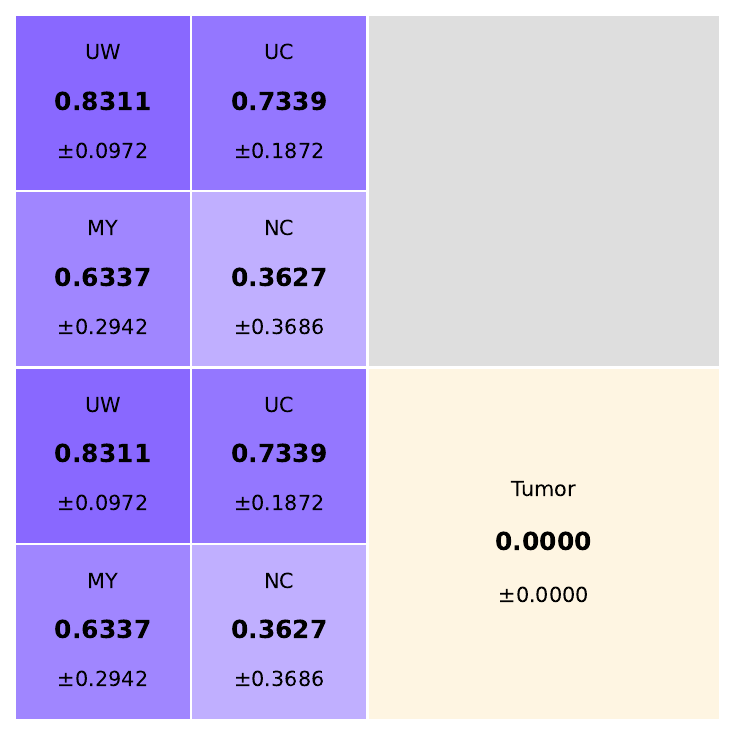} &
\includegraphics[width=0.23\textwidth]{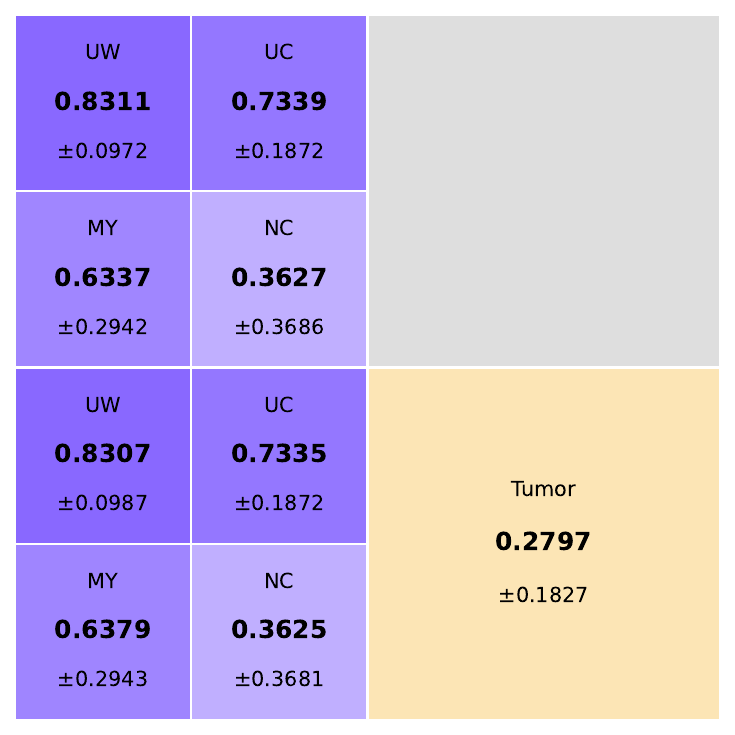} &
\includegraphics[width=0.23\textwidth]{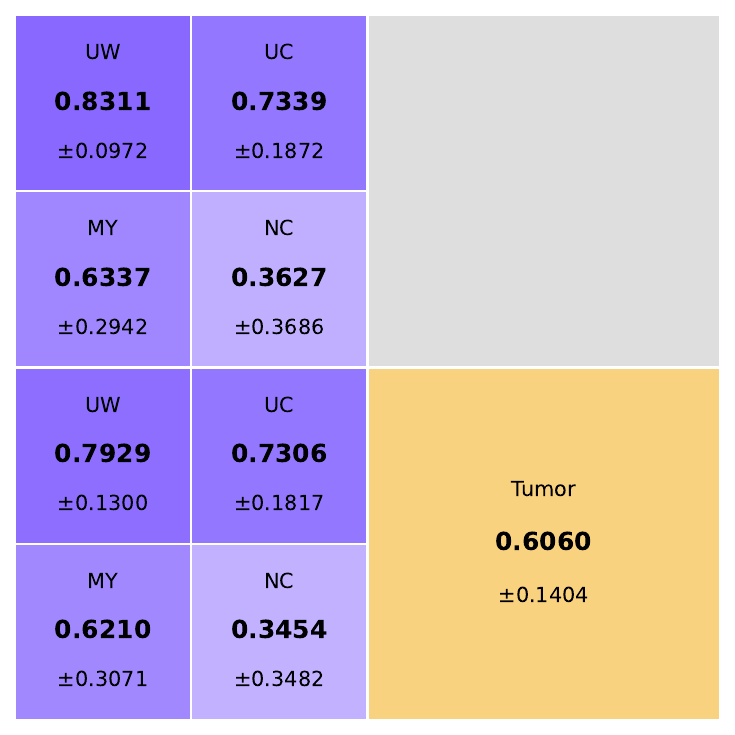} &
\includegraphics[width=0.23\textwidth]{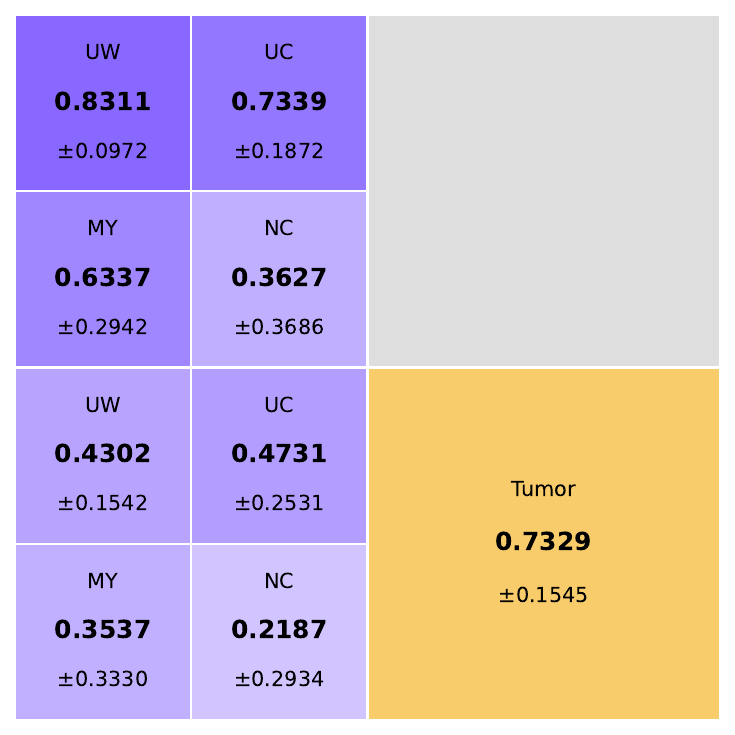} \\
(a) & (b) & (c) & (d)
\end{tabular}

\vspace{1.5mm}

\begin{tabular}{ccc}
\includegraphics[width=0.23\textwidth]{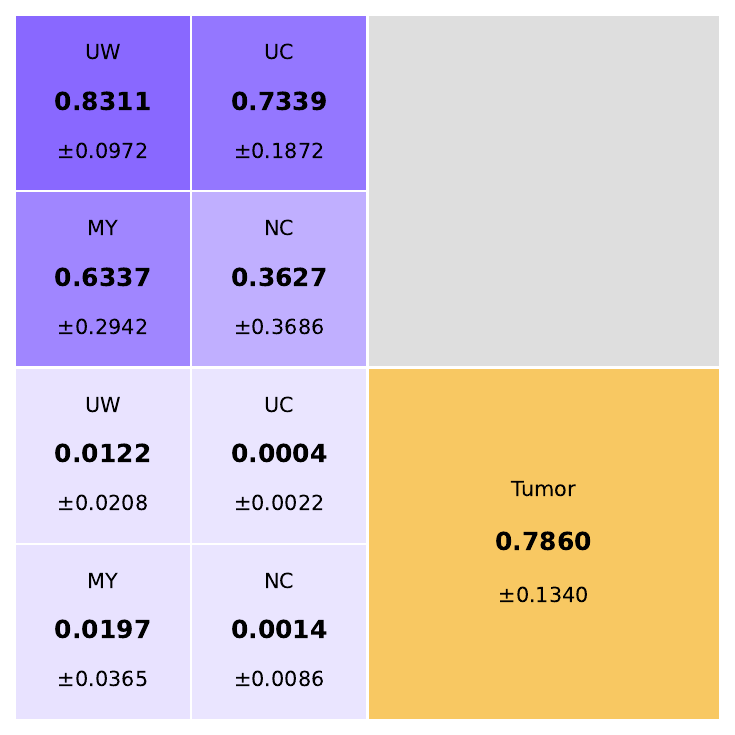} &
\includegraphics[width=0.23\textwidth]{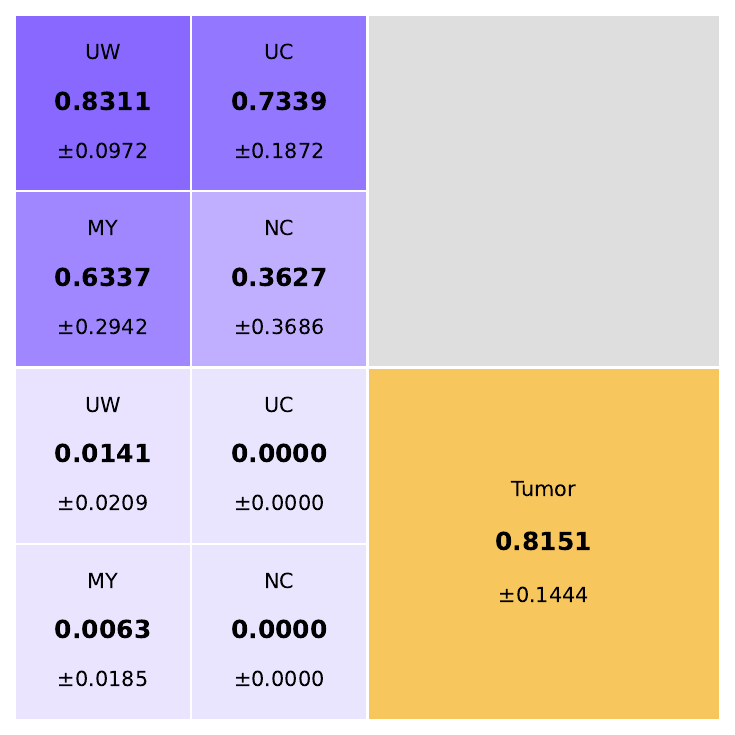} &
\includegraphics[width=0.23\textwidth]{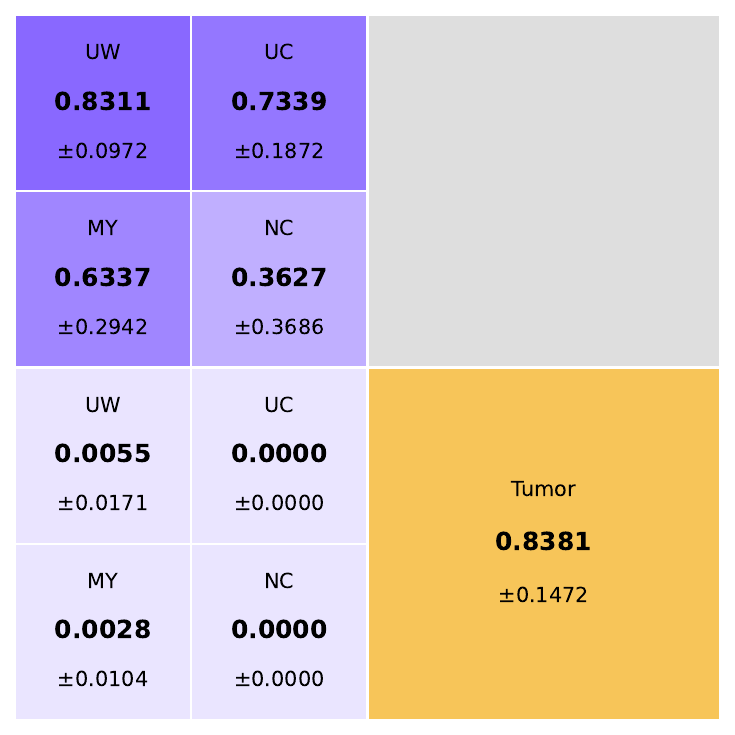} \\
(e) & (f) & (g)
\end{tabular}

\caption{Progressive depth-constrained adaptation during UMD $\rightarrow$ ECPC-IDS. (a) Fully frozen backbone; (b) bottleneck only; (c) $E_5$ + bottleneck + $D_5$; (d) $E_4$--$E_5$ + bottleneck + $D_4$--$D_5$; (e) $E_3$--$E_5$ + bottleneck + $D_3$--$D_5$; (f) $E_2$--$E_5$ + bottleneck + $D_2$--$D_5$; (g) $E_1$--$E_5$ + bottleneck + $D_1$--$D_5$.}
\label{fig:block_heatmaps_progressive}
\end{figure*}

\begin{figure}[t]
\centering
\setlength{\tabcolsep}{1pt}
\renewcommand{\arraystretch}{0.9}
\scriptsize

\begin{tabular}{cccc}
\textbf{(a)} & \textbf{(b)} & \textbf{(c)} & \textbf{(d)} \\
\includegraphics[width=0.23\linewidth]{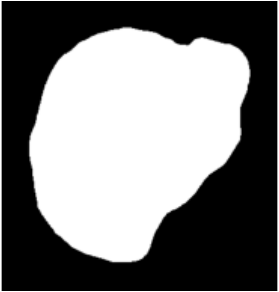} &
\includegraphics[width=0.23\linewidth]{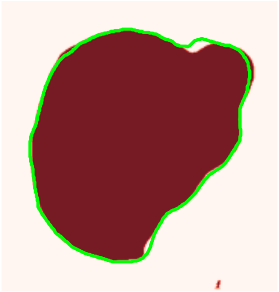} &
\includegraphics[width=0.23\linewidth]{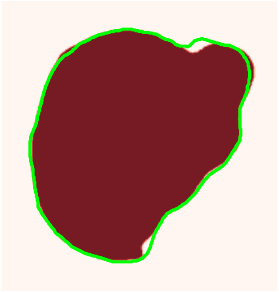} &
\includegraphics[width=0.23\linewidth]{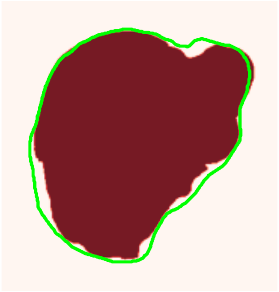} \\[1.5mm]

\textbf{(e)} & \textbf{(f)} & \textbf{(g)} & \textbf{(h)} \\
\includegraphics[width=0.23\linewidth]{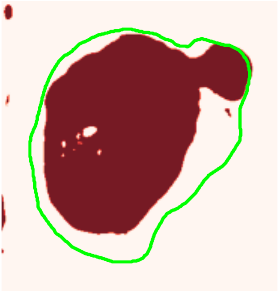} &
\includegraphics[width=0.23\linewidth]{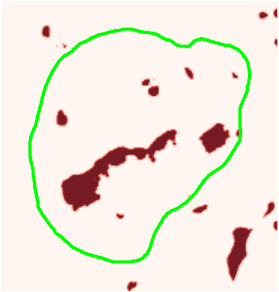} &
\includegraphics[width=0.23\linewidth]{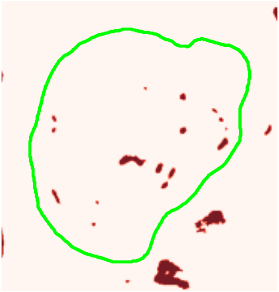} &
\includegraphics[width=0.23\linewidth]{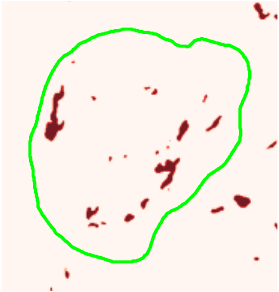}
\end{tabular} 

\caption{Qualitative visualization of forgetting on a representative UMD case under depth-constrained adaptation. Green contours denote ground-truth masks and red regions denote predictions. (a) Ground truth; (b) fully frozen backbone; (c) bottleneck only; (d) $E_5$ + bottleneck + $D_5$; (e) $E_4$--$E_5$ + bottleneck + $D_4$--$D_5$; (f) $E_3$--$E_5$ + bottleneck + $D_3$--$D_5$; (g) $E_2$--$E_5$ + bottleneck + $D_2$--$D_5$; (h) $E_1$--$E_5$ + bottleneck + $D_1$--$D_5$.}
\label{fig:depth_qualitative}
\end{figure}

This section provides additional results for the progressive UMD $\rightarrow$ ECPC-IDS experiment reported in Section~\ref{sec:depth_results}. Figure~\ref{fig:block_heatmaps_progressive} shows the complete task-wise performance progression, Table~\ref{tab:progressive_forgetting} reports forgetting for individual UMD structures, and Figure~\ref{fig:depth_qualitative} provides a representative qualitative example.

% --- TABLE (Single-column ML4H layout) ---
\begin{table}[t]
\centering
\caption{UMD per-label forgetting under progressive block unfreezing during UMD $\rightarrow$ ECPC-IDS. Lower values indicate better retention.}
\label{tab:progressive_forgetting}
\resizebox{\linewidth}{!}{%
\begin{tabular}{lcccc}
\toprule
\textbf{Trainable Region} & \textbf{UW} & \textbf{UC} & \textbf{MY} & \textbf{NC} \\
\midrule
Frozen                                   & 0.0000 & 0.0000 & 0.0000  & 0.0000 \\
Bottleneck                               & 0.0004 & 0.0004 & -0.0042 & 0.0002 \\
$E_5$ + Bottleneck + $D_5$               & 0.0382 & 0.0034 & 0.0127  & 0.0173 \\
$E_4$--$E_5$ + Bottleneck + $D_4$--$D_5$ & 0.4009 & 0.2608 & 0.2800  & 0.1439 \\
$E_3$--$E_5$ + Bottleneck + $D_3$--$D_5$ & 0.8189 & 0.7336 & 0.6140  & 0.3612 \\
$E_2$--$E_5$ + Bottleneck + $D_2$--$D_5$ & 0.8170 & 0.7339 & 0.6274  & 0.3627 \\
$E_1$--$E_5$ + Bottleneck + $D_1$--$D_5$ & 0.8256 & 0.7339 & 0.6309  & 0.3627 \\
\bottomrule
\end{tabular}%
}
\end{table}

% --- FIGURE (Single-column ML4H layout) ---

Across these analyses, the same depth-wise pattern observed in the aggregate results is apparent. Bottleneck-adjacent adaptation largely preserves previous-task predictions, while forgetting increases sharply as the trainable region expands through more input- and output-proximal stages. The class-wise results show that this degradation occurs across all UMD structures rather than being driven by a single label, while the qualitative example illustrates the corresponding loss of previous-task segmentation performance.

%========================================================================================%

\section{Parameter-Displacement and Learning-Rate Details}
\label{sec:app_displacement}

This section provides supporting details for the RMS-targeted and scale-aware
experiments. All displacement values exclude task-specific heads.

\subsection{Approximate RMS-Targeting Calculation}
\label{sec:app_rms_targeting}

The bottleneck RMS, $0.004424$, defines a target range of
$[0.003982,0.004866]$ ($\pm10\%$). The middle-region RMS, $0.004859$, already
lies within this range. For the broad region, we estimate the initial learning
rate by scaling its shared-LR value according to
\[
\eta_{\mathrm{broad}} \approx 0.01
\frac{0.004424}{0.005512}=0.00803.
\]
We therefore use an initial learning rate of $0.008$ for the broad run. This is
an approximate final-checkpoint control; RMS is measured after training and is
not constrained during optimization.

\subsection{Middle-Region Block Displacement}
\label{sec:app_middle_displacement}

Table~\ref{tab:middle_block_displacement} compares the same trainable region
$\mathcal{R}_4$ under shared and scale-aware learning rates. Scale-aware
calibration reduces movement in the outer trainable pair, $E_4/D_4$, while
leaving bottleneck RMS almost unchanged. Consequently, the combined
$E_4/D_4$ share falls from $55.9\%$ to $12.2\%$, and the bottleneck share rises
from $12.2\%$ to $55.1\%$.

\begin{table*}[t]
\floatconts
  {tab:middle_block_displacement}
  {\caption{Per-block displacement within the identical middle region
  $\mathcal{R}_4$. The squared-displacement shares sum to 100\% within each
  condition.}}
  {%
  \small
  \begin{tabular*}{\textwidth}{@{\extracolsep{\fill}}lcccccc}
  \toprule
  & \multicolumn{3}{c}{\textbf{Shared LR}} &
    \multicolumn{3}{c}{\textbf{Scale-aware LR}} \\
  \cmidrule(lr){2-4}\cmidrule(lr){5-7}
  \textbf{Block} & \textbf{RMS} & \textbf{Relative} & \textbf{$q_b$} &
  \textbf{RMS} & \textbf{Relative} & \textbf{$q_b$} \\
  \midrule
  $E_4$ & 0.00631 & 40.8\% & 24.8\% & 0.00110 & 7.1\% & 3.5\% \\
  $E_5$ & 0.00457 & 34.0\% & 14.5\% & 0.00221 & 16.4\% & 15.5\% \\
  $B$   & 0.00420 & 32.9\% & 12.2\% & 0.00417 & 32.6\% & 55.1\% \\
  $D_5$ & 0.00400 & 33.5\% & 17.4\% & 0.00186 & 15.5\% & 17.2\% \\
  $D_4$ & 0.00522 & 41.0\% & 31.1\% & 0.00129 & 10.1\% & 8.7\% \\
  \bottomrule
  \end{tabular*}
  }
\end{table*}

\subsection{Full-Backbone Scale-Aware Displacement}
\label{sec:app_full_control}

Table~\ref{tab:full_block_displacement} reports the scale-aware result when all
13 backbone blocks are trainable. Its aggregate RMS is $0.002222$, but UMD
forgetting is $0.6404$ and ECPC-IDS DSC is $0.8670$. Thus, reducing average
movement does not preserve the previous task when the full backbone is exposed.

\begin{table}[t]
\centering
\caption{Full-backbone scale-aware displacement. $q_b$ is the share of total
squared backbone displacement; percentages may not sum to 100 due to rounding.}
\label{tab:full_block_displacement}
\small
\begin{tabular*}{\columnwidth}{@{\extracolsep{\fill}}lccc}
\toprule
\textbf{Block} & \textbf{RMS} & \textbf{Relative} & \textbf{$q_b$} \\
\midrule
$E_0$ & 0.03061 & 22.2\% & 4.3\% \\
$E_1$ & 0.00983 & 12.7\% & 2.5\% \\
$E_2$ & 0.00354 & 6.9\% & 1.3\% \\
$E_3$ & 0.00121 & 5.9\% & 1.8\% \\
$E_4$ & 0.00079 & 5.1\% & 1.4\% \\
$E_5$ & 0.00205 & 15.3\% & 10.6\% \\
$B$ & 0.00415 & 32.5\% & 43.4\% \\
$D_5$ & 0.00175 & 14.7\% & 12.2\% \\
$D_4$ & 0.00084 & 6.6\% & 2.9\% \\
$D_3$ & 0.00104 & 6.9\% & 2.9\% \\
$D_2$ & 0.00299 & 12.3\% & 6.0\% \\
$D_1$ & 0.00867 & 15.1\% & 4.9\% \\
$D_0$ & 0.01898 & 21.8\% & 5.9\% \\
\bottomrule
\end{tabular*}
\end{table}

\subsection{Block-Specific Learning-Rate Multipliers}
\label{sec:app_lr_multipliers}

Tables~\ref{tab:lr_multipliers} and~\ref{tab:full_lr_multipliers} report the
static multipliers obtained from 20 gradient-only calibration batches. The
bottleneck is the reference with multiplier $1$. Each multiplier remains fixed
while the common epoch-wise schedule decays. Calibration rescales the learning
rates but does not force the blocks to have equal final displacement.

\begin{table*}[t]
\floatconts
  {tab:lr_multipliers}
  {\caption{Static LR multipliers $\alpha_b$ obtained from the initial
  relative-gradient calibration. A dash denotes a frozen block.}}
  {%
  \small
  \begin{tabular*}{\textwidth}{@{\extracolsep{\fill}}lccccccc}
  \toprule
  \textbf{Region} & $E_3$ & $E_4$ & $E_5$ & $B$ & $D_5$ & $D_4$ & $D_3$ \\
  \midrule
  $\mathcal{R}_5$ & -- & -- & 0.462 & 1.000 & 0.436 & -- & -- \\
  $\mathcal{R}_4$ & -- & 0.142 & 0.443 & 1.000 & 0.443 & 0.179 & -- \\
  $\mathcal{R}_3$ & 0.100 & 0.118 & 0.446 & 1.000 & 0.412 & 0.139 & 0.117 \\
  \bottomrule
  \end{tabular*}
  }
\end{table*}

\begin{table}[t]
\centering
\caption{Calibrated LR multipliers for the full backbone. $E_3$ reaches the
lower clipping bound. The task-specific head multiplier is 1.000.}
\label{tab:full_lr_multipliers}
\small
\begin{tabular*}{\columnwidth}{@{\extracolsep{\fill}}lc}
\toprule
\textbf{Block} & \textbf{$\alpha_b$} \\
\midrule
$E_0$ & 0.211 \\
$E_1$ & 0.184 \\
$E_2$ & 0.102 \\
$E_3$ & 0.100 \\
$E_4$ & 0.111 \\
$E_5$ & 0.418 \\
$B$ & 1.000 \\
$D_5$ & 0.398 \\
$D_4$ & 0.147 \\
$D_3$ & 0.121 \\
$D_2$ & 0.183 \\
$D_1$ & 0.213 \\
$D_0$ & 0.248 \\
\bottomrule
\end{tabular*}
\end{table}

\section{Uncertainty Analyses}
\label{sec:app_uncertainty}

\subsection{Additional Training-Seed Results}
\label{sec:app_seed_results}

We repeat the bottleneck, middle, and RMS-targeted broad configurations, as
well as the shared and scale-aware learning-rate configurations, over three
adaptation seeds (12345, 23456, 34567).

\begin{table}[t]
\centering
\caption{Individual adaptation-seed results for the approximate-RMS comparison.
Negative forgetting denotes a small improvement in UMD DSC.}
\label{tab:seed_results}
\resizebox{\columnwidth}{!}{%
\begin{tabular}{llcc}
\toprule
\textbf{Configuration} & \textbf{Seed} & \textbf{UMD Forgetting $\downarrow$} & \textbf{ECPC DSC $\uparrow$} \\
\midrule
Bottleneck & 12345 & $-0.0008$ & 0.2797 \\
 & 23456 & 0.0008 & 0.0000 \\
 & 34567 & $-0.0000$ & 0.3782 \\
\midrule
Middle ($\mathcal{R}_4$) & 12345 & 0.2714 & 0.7329 \\
 & 23456 & 0.4073 & 0.7515 \\
 & 34567 & 0.4100 & 0.7367 \\
\midrule
Broad, RMS-targeted ($\mathcal{R}_1$) & 12345 & 0.6384 & 0.8342 \\
 & 23456 & 0.6403 & 0.8454 \\
 & 34567 & 0.6404 & 0.8449 \\
\bottomrule
\end{tabular}}
\end{table}

The three seeds preserve the same ordering: bottleneck adaptation retains UMD,
middle adaptation gives an intermediate trade-off, and broad adaptation nearly
completely forgets UMD. Mean forgetting $\pm$ sample SD is
$0.0000\pm0.0008$, $0.3629\pm0.0793$, and $0.6397\pm0.0011$, respectively.

Table~\ref{tab:scale_aware_seeds} reports per-seed results for the shared and
scale-aware learning-rate experiments. The ordering is preserved across all
seeds: forgetting rises sharply beyond $\mathcal{R}_5$, and scale-aware rates
substantially reduce forgetting at $\mathcal{R}_4$ and $\mathcal{R}_3$ with
only small reductions in current-task Dice.

\begin{table}[t]
\centering
\caption{Per-seed results for shared and scale-aware learning rates
(UMD $\rightarrow$ ECPC-IDS). Scale-aware rates are calibrated independently
per seed. The $\mathcal{R}_5$ shared-LR Dice reflects corrected PET/CT
evaluation preprocessing.}
\label{tab:scale_aware_seeds}
\resizebox{\columnwidth}{!}{%
\begin{tabular}{llccc}
\toprule
\textbf{Region} & \textbf{LR} & \textbf{Seed} &
\textbf{UMD Forgetting $\downarrow$} & \textbf{ECPC DSC $\uparrow$} \\
\midrule
$\mathcal{R}_5$ & Shared & 12345 & 0.0179 & 0.6060 \\
 & & 23456 & 0.0226 & 0.5978 \\
 & & 34567 & 0.0196 & 0.6044 \\
\cmidrule{2-5}
 & Scale-aware & 12345 & $-$0.0011 & 0.6025 \\
 & & 23456 & 0.0058 & 0.5921 \\
 & & 34567 & $-$0.0034 & 0.6063 \\
\midrule
$\mathcal{R}_4$ & Shared & 12345 & 0.2714 & 0.7329 \\
 & & 23456 & 0.4073 & 0.7515 \\
 & & 34567 & 0.4100 & 0.7367 \\
\cmidrule{2-5}
 & Scale-aware & 12345 & 0.0252 & 0.7183 \\
 & & 23456 & 0.0542 & 0.7292 \\
 & & 34567 & 0.0554 & 0.7251 \\
\midrule
$\mathcal{R}_3$ & Shared & 12345 & 0.6319 & 0.7860 \\
 & & 23456 & 0.6253 & 0.7858 \\
 & & 34567 & 0.6148 & 0.7839 \\
\cmidrule{2-5}
 & Scale-aware & 12345 & 0.1652 & 0.7715 \\
 & & 23456 & 0.2168 & 0.7730 \\
 & & 34567 & 0.2155 & 0.7736 \\
\midrule
$\mathcal{R}_0$ & Shared & 12345 & 0.6404 & 0.8658 \\
 & (all seeds) & & $0.640\pm0.000$ & $0.865\pm0.004$ \\
\cmidrule{2-5}
 & Scale-aware & 12345 & 0.6404 & 0.8670 \\
 & (all seeds) & & $0.640\pm0.000$ & $0.863\pm0.004$ \\
\bottomrule
\end{tabular}}
\end{table}

\subsection{Held-Out-Case Bootstrap Intervals}
\label{sec:app_bootstrap}

Table~\ref{tab:bootstrap_intervals} reports 95\% confidence intervals from
10,000 held-out-case bootstrap resamples. UMD forgetting uses paired resampling
of the same cases before and after adaptation. These intervals measure
finite-cohort uncertainty, not variation across training seeds. The
UMD$\rightarrow$EndoMRI intervals are descriptive because the T1FS sequence
was chosen in an exploratory comparison that included the held-out cohort.

\begin{table*}[t]
\floatconts
  {tab:bootstrap_intervals}
  {\caption{Point estimates and 95\% held-out-case bootstrap confidence
  intervals. These intervals do not measure training-seed variability.}}
  {%
  \small
  \begin{tabular*}{\textwidth}{@{\extracolsep{\fill}}lllcc}
  \toprule
  \textbf{Transition} & \textbf{Region} & \textbf{LR allocation} &
  \textbf{Forgetting [95\% CI]} & \textbf{Current-task Dice [95\% CI]} \\
  \midrule
  UMD$\rightarrow$ECPC & $\mathcal{R}_5$ & Shared
    & 0.0179 [0.0012, 0.0344] & 0.6060 [0.5494, 0.6542] \\
  UMD$\rightarrow$ECPC & $\mathcal{R}_5$ & Scale-aware
    & $-0.0011$ [$-0.0153$, 0.0115] & 0.6025 [0.5445, 0.6507] \\
  UMD$\rightarrow$ECPC & $\mathcal{R}_4$ & Shared
    & 0.2714 [0.2390, 0.3043] & 0.7329 [0.6694, 0.7833] \\
  UMD$\rightarrow$ECPC & $\mathcal{R}_4$ & Scale-aware
    & 0.0252 [0.0108, 0.0412] & 0.7183 [0.6578, 0.7651] \\
  UMD$\rightarrow$ECPC & $\mathcal{R}_3$ & Shared
    & 0.6319 [0.5918, 0.6693] & 0.7860 [0.7276, 0.8275] \\
  UMD$\rightarrow$ECPC & $\mathcal{R}_3$ & Scale-aware
    & 0.1652 [0.1373, 0.1933] & 0.7715 [0.7108, 0.8176] \\
  \midrule
  UMD$\rightarrow$Endo & $B$ & Shared
    & 0.0592 [0.0360, 0.0836] & 0.0985 [0.0776, 0.1204] \\
  UMD$\rightarrow$Endo & $\mathcal{R}_4$ & Shared
    & 0.6380 [0.5982, 0.6777] & 0.2435 [0.1919, 0.2942] \\
  UMD$\rightarrow$Endo & $\mathcal{R}_1$ & LR 0.008
    & 0.6362 [0.5968, 0.6742] & 0.3019 [0.2353, 0.3669] \\
  \bottomrule
  \end{tabular*}
  }
\end{table*}

\end{document}